\documentclass[runningheads]{llncs}

\usepackage{eccv}

\usepackage{eccvabbrv}

\usepackage{graphicx}

\usepackage{url}            % simple URL typesetting
\usepackage{booktabs}       % professional-quality tables
\usepackage{amsfonts}       % blackboard math symbols
\usepackage{nicefrac}       % compact symbols for 1/2, etc.
\usepackage{microtype}      % microtypography
\usepackage{xcolor}         % colors
\usepackage[ruled,linesnumbered]{algorithm2e}
\usepackage{arydshln}

\usepackage{enumitem}
\usepackage{bbm}
\usepackage{wrapfig}

\usepackage[dvipsnames]{xcolor}
\usepackage{multicol}
\usepackage{colortbl}
\usepackage{multirow}
\usepackage{pifont}
\usepackage{makecell}

\usepackage[accsupp]{axessibility}  % Improves PDF readability for those with disabilities.

\usepackage{hyperref}

\usepackage{orcidlink}
\usepackage{marvosym}

\begin{document}

\title{Dense Multimodal Alignment for Open-Vocabulary 3D Scene Understanding} 

\titlerunning{Dense Multimodal Alignment}

\author{Ruihuang Li\inst{1} \and
Zhengqiang Zhang\inst{1} \and
Chenhang He\inst{1} \and
Zhiyuan Ma\inst{1} \and \\
Vishal M. Patel\inst{2} \and
Lei Zhang\inst{1}\textsuperscript{(\Email)} 
%\thanks{Corresponding author}
}

\authorrunning{R.~Li, Z.~Zhang, et al.}

\institute{Hong Kong Polytechnic University \and
Joins Hopkins University\\
\email{\{csrhli, cslzhang\}@comp.polyu.edu.hk, vpatel36@jhu.edu} \\
\url{https://github.com/lslrh/DMA}}

\maketitle

\begin{abstract}
Recent vision-language pre-training models have exhibited remarkable generalization ability in zero-shot recognition tasks. Previous open-vocabulary 3D scene understanding methods mostly focus on training 3D models using either image or text supervision while neglecting the collective strength of all modalities. In this work, we propose a Dense Multimodal Alignment (DMA) framework to densely co-embed different modalities into a common space for maximizing their synergistic benefits. Instead of extracting coarse view- or region-level text prompts, we leverage large vision-language models to extract complete category information and scalable scene descriptions to build the text modality, and take image modality as the bridge to build dense point-pixel-text associations. Besides, in order to enhance the generalization ability of the 2D model for downstream 3D tasks without compromising the open-vocabulary capability, we employ a dual-path integration approach to combine frozen CLIP visual features and learnable mask features. Extensive experiments show that our DMA method produces highly competitive open-vocabulary segmentation performance on various indoor and outdoor tasks. 
\end{abstract}
%However, their applications to 3D dense prediction tasks often encounter the difficulties of limited high-quality and densely-annotated 3D data.

\section{Introduction}
\label{sec:intro}
3D scene understanding, which aims to achieve accurate comprehension of objects as well as their attributes and relationships within a scene, has gained significant attention in recent years due to its popular applications in autonomous driving~\cite{li2022deepfusion}, virtual reality (VR)~\cite{armeni20163d,misra2021end,vu2022softgroup} and robot navigation~\cite{cadena2016multi}, \etc. However, the annotation of large-scale 3D data is very costly ~\cite{chang2015shapenet,dai2017scannet}, impeding the training of generalizable models for open-vocabulary scene understanding. Though many existing methods~\cite{choy20194d,hu2021vmnet,zhu2021cylindrical,schult2020dualconvmesh,nekrasov2021mix3d,li2022class,li2023dynamask,li2023sim,chen2023fpr} have achieved significant advancements in recognizing closed-set categories for specific tasks, they fail to identify novel categories and other types of queries~\cite{peng2023openscene} without 3D supervision, hindering the application of existing 3D scene understanding methods to real-world settings, where the number of possible classes is unlimited. 

%\begin{figure}
%	\centering 
%	\includegraphics[scale=0.8]{overview.jpg}\\
%	\vspace{-0.5em}
%	\caption{Training pipelines of different open-vocabulary 3D scene understanding methods. (a) Distilling the knowledge from pre-trained 2D encoders to 3D representations. (b) Aligning 3D features to text embeddings. (c) Our proposed DMA method, which constructs dense correspondences between 3D and other modalities, and embeds them into a common latent space.  }
%	\label{DMA:overview}
%	\vspace{-2mm}
%\end{figure}

In contrast to the limited 3D data,  modalities such as images and texts are more abundantly available. Existing pre-trained multimodal models, such as CLIP~\cite{radford2021learning} and ALIGN~\cite{jia2021scaling}, have shown impressive zero-shot recognition ability by training on large-scale noisy image-text pairs, and have been successfully adapted for open-vocabulary classification~\cite{xue2023ulip,xue2023ulip2}, detection~\cite{lu2023open,cao2024coda} and segmentation tasks~\cite{liang2023open,xu2023open,takmaz2023openmask3d}. Based on these observations, researchers have attempted to use image or natural language modalities to provide supervisory signals for learning 3D representations~\cite{yang2023regionplc,ding2023pla,peng2023openscene,liu2023weakly}. Some methods use fixed 2D features as supervision and distill the knowledge from either the pre-trained 2D encoder of CLIP~\cite{liu2023weakly} or 2D open-vocabulary segmentation (OVSeg) models~\cite{peng2023openscene} into 3D representations (NeRF or point clouds). However, they overlook the fact that 3D models can in turn enhance 2D models by leveraging the strong 3D structural information. Besides, the 2D OVSeg models compromise their open-vocabulary ability since they are primarily fine-tuned on in-vocabulary datasets. There are also some methods that directly align 3D features to semantic captions~\cite{yang2023regionplc,ding2023pla,rozenberszki2022language}. However, they only capture coarse image- or region-level descriptions without establishing dense point-to-text correspondences or exploiting image features that involve rich semantic contexts and more variations. Though some methods~\cite{xue2023ulip,xue2023ulip2} simultaneously leverage visual and textual supervisions, they only conduct coarse multimodal alignment for object-level point cloud classification.  

% there exist several notable limitations that need to be addressed. First, directly applying the features of frozen CLIP to dense prediction tasks often yields suboptimal results in practice due to the domain shift from full images to local image regions. are primarily fine-tuned on datasets with a limited vocabulary~\cite{li2022languagedriven,ghiasi2022scaling}, thereby compromising their ability to identify open-vocabulary classes. Besides, these methods utilize 2D features as supervision, thereby resulting in a bottleneck in the performance of 3D models, which is determined by the performance of the 2D models. 

In order to leverage the synergistic benefits of multiple modalities for dense prediction tasks, we propose a dense multimodal alignment (DMA) strategy to co-embed 3D points, image pixels, and text strings into a shared latent space. To build dense associations across different modalities, the primary bottleneck is \textit{how to obtain rich and reliable text descriptions without relying on manual labeling}. To this end, we generate two types of prompts using large Vision-Language Models (VLMs). Firstly, we employ the tagging model such as  RAM~\cite{zhang2023recognize} to detect as many categories as possible from an image, ensuring alignment with \textbf{complete} semantic patterns. Considering that category names might not provide sufficient details and contextual information, we incorporate Multimodal Large Language Models (MLLM) such as LLaVA~\cite{liu2023llava} to generate linguistically expressible scene descriptions, thereby enhancing the \textbf{scalability} of text queries. In addition, we use the GPT to filter out the noise in the generated texts for improving the \textbf{reliability}. As a result, we establish a highly scalable and informative text modality, enhancing the overall understanding of 3D scenes.

As for the image modality, we adopt a \textbf{dual-path integration} strategy to extract robust 2D features as supervision. Specifically, we employ the FC-CLIP~\cite{yu2023fcclip} as the feature extractor. On one hand, we fix its CLIP visual encoder to maintain the open-world recognition ability. On the other hand, by fine-tuning its mask head, we  \textit{incorporate 3D structural priors into 2D features}, better adapting the model to 3D dense tasks. Then we build triplets of points, pixels, and their corresponding texts by taking image modality as the bridge. Given the generated triplets of different modalities and their dense correspondences, we finally adopt the \textbf{mutually inclusive loss} function to align multiple modalities. In this way, we can effectively unleash the potential of existing foundation VLMs and maximize the complementary effects of multiple modalities. 

In summary, (1) we first present a dense multimodal alignment framework, which establishes dense correspondences among points, pixels and texts, to learn robust 3D representations for open-vocabulary 3D scene understanding. (2) To generate complete and scalable language modality without relying on manual annotations, we leverage a tagging model and an MLLM to extract category information and scene descriptions, respectively. (3) Finally, to improve the segmentation ability without compromising the open-vocabulary ability, we integrate 3D priors into 2D features by fine-tuning the 2D mask head with the backbone frozen. Extensive experiments demonstrate the outstanding open-world 3D segmentation ability of our DMA model on various indoor and outdoor tasks. 

\section{Related Work}
{\bf Open-Vocabulary 3D Scene Understanding.}
3D scene understanding is a popular research topic in computer vision. Most previous methods~\cite{choy20194d,han2020occuseg,hu2021bidirectional,hu2021vmnet,zhu2021cylindrical} focus on training models on manually labeled close-set categories, and have yielded promising performance on popular 3D benchmarks~\cite{dai2017scannet,caesar2020nuscenes}. However, most of these methods are designed for a specific task, such as object classification~\cite{wu20153d}, detection~\cite{chen2020scanrefer}, semantic/instance segmentation~\cite{zhu2021cylindrical,choy20194d,han2020occuseg}, and they cannot identify novel categories, restricting their applications to real-world settings. To overcome this limitation, recent works have been focused on the open-vocabulary scene understanding problem~\cite{liu2023weakly}. Rozenberszki~\etal~\cite{rozenberszki2022language} proposed a language-driven pre-training method to enforce 3D feature to be close to text embeddings, and finetune the 3D encoder with ground-truth annotations. PLA~\cite{ding2023pla} and RegionPLC~\cite{yang2023regionplc} explicitly associate 3D points with image- and region-level image captions, respectively. However, existing image captioning models can only identify sparse and salient objects while missing other important categories. Besides, textual signals lack variations and contexts, making them insufficient for dense prediction tasks. Some methods~\cite{peng2023openscene,liu2023weakly} distill knowledge from large-scale pre-trained 2D models, such as image-text contrastive learning models~\cite{radford2021learning} and open-vocabulary segmentation models~\cite{liang2023open,xu2023open,ghiasi2022scaling,yu2023fcclip}. However, the performance of pre-trained models drops a lot on the downstream datasets due to the large domain shift. These methods also overlook the fact that 3D models can in turn enhance 2D models by leveraging the strong structural information inherent in 3D data. 

\begin{figure*}[!t]
	\centering 
	\includegraphics[scale=0.5]{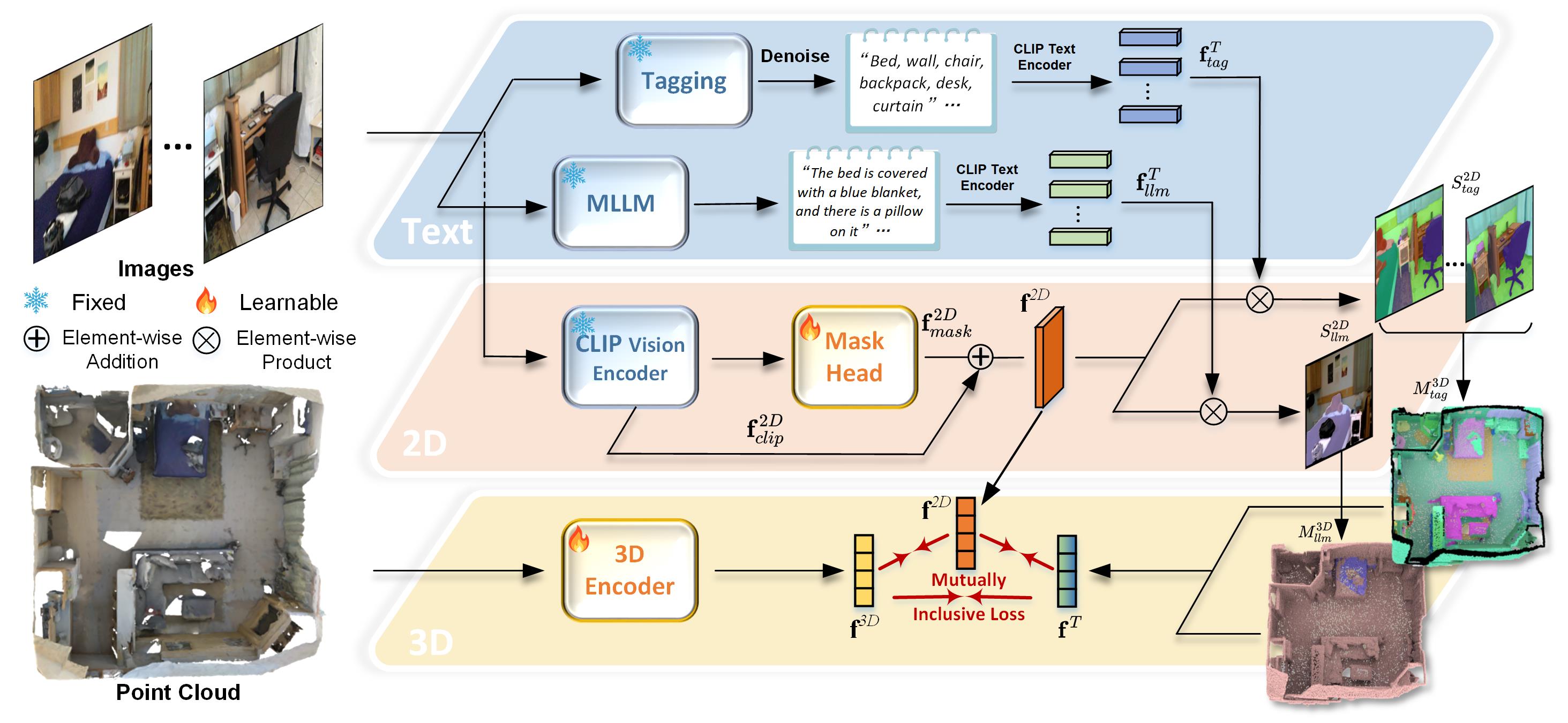}\\ \vspace{-0.2cm}
	\caption{Framework of our proposed Dense Multimodal Alignment (DMA) method. We generate comprehensive language modality data by leveraging a tagging model and an MLLM. As for 2D modality, we fix the CLIP visual backbone ${\bf f}^{2D}_{clip}$ but finetune the mask head ${\bf f}^{2D}_{mask}$ for better adaptation to downstream 3D tasks without compromising the open-vocabulary ability. Then the dense correspondences between pixels ${\bf f}^{2D}$ and texts ${\bf f}^{T}_{tag}$/${\bf f}^{T}_{llm}$ can be built by computing their feature similarities, resulting in semantic score maps $S^{2D}_{tag}$/$S^{2D}_{llm}$. By taking image modality as the bridge, we back-project text labels to each point and obtain the 3D label maps $M^{3D}_{tag}$/$M^{3D}_{llm}$. Finally, we co-embed point ${\bf f}^{3D}$, pixel ${\bf f}^{2D}$, and text embeddings ${\bf f}^{T}$ into a common space to learn a robust 3D representation by optimizing the mutually inclusive loss function.}
	\label{DMA:framework} 
	\vspace{-0.3cm}
\end{figure*}
{\bf Vision-Language Foundation Models.} Recent vision-language foundation models have exhibited remarkable generalization ability on zero-shot prediction tasks. Segment Anything Model (SAM)~\cite{Kirillov_2023_ICCV} leads a new trend of universal image segmentation and exhibits promising results on diverse downstream tasks. Recognize Anything Model (RAM)~\cite{zhang2023recognize} presents a novel paradigm for image tagging (multi-label classification) by leveraging large-scale image-text pairs for training without manual annotations. The recent success of ChatGPT and GPT4 have stimulated tremendous interests in developing multimodal large language models (LLMs). LLaVA~\cite{liu2023llava} is an early exploration to apply LLMs to the multimodal fields by connecting a vision encoder to LLM for general-purpose visual and language understanding. The recent open-vocabulary methods~\cite{xu2023open,liang2023open,yu2023fcclip} shed lights on the direct use of pre-trained foundation models for handling different visual tasks. ODISE~\cite{xu2023open} explores the potential ability of pretrained text-to-image diffusion models~\cite{rombach2022high} for open-vocabulary panoptic segmentation. FC-CLIP~\cite{yu2023fcclip} utilizes a shared frozen convolutional CLIP backbone to maintain the ability of open-vocabulary classification without compromising accuracy.

\vspace{-0.2cm}
\section{Method}
As illustrated in Fig.~\ref{DMA:framework}, we propose a dense multimodal alignment (DMA) framework for open-vocabulary 3D scene understanding, where we construct dense correspondences across 2D image pixels, 3D points and 1D texts, and embed them into a common latent space. In this section, we will elaborate the construction of text and image modalities, and explain how we associate and align them in a dense manner.

\vspace{-0.2cm} 
\subsection{Comprehensive Text Modality Generation}
Learning a robust 3D model that is generalizable to open vocabularies is challenging since it is unclear how to acquire the dense text labels for point clouds. Although well-trained human annotators could potentially provide detailed language descriptions of 3D scenes, such a method is costly and lacks scalability. To overcome this limitation, we leverage a tagging model and an MLLM to extract \textbf{complete} category information and \textbf{scalable} scene descriptions, respectively. 

\begin{figure}[!t]
	\centering 
	\hspace{-1.0cm}
	\includegraphics[scale=0.68]{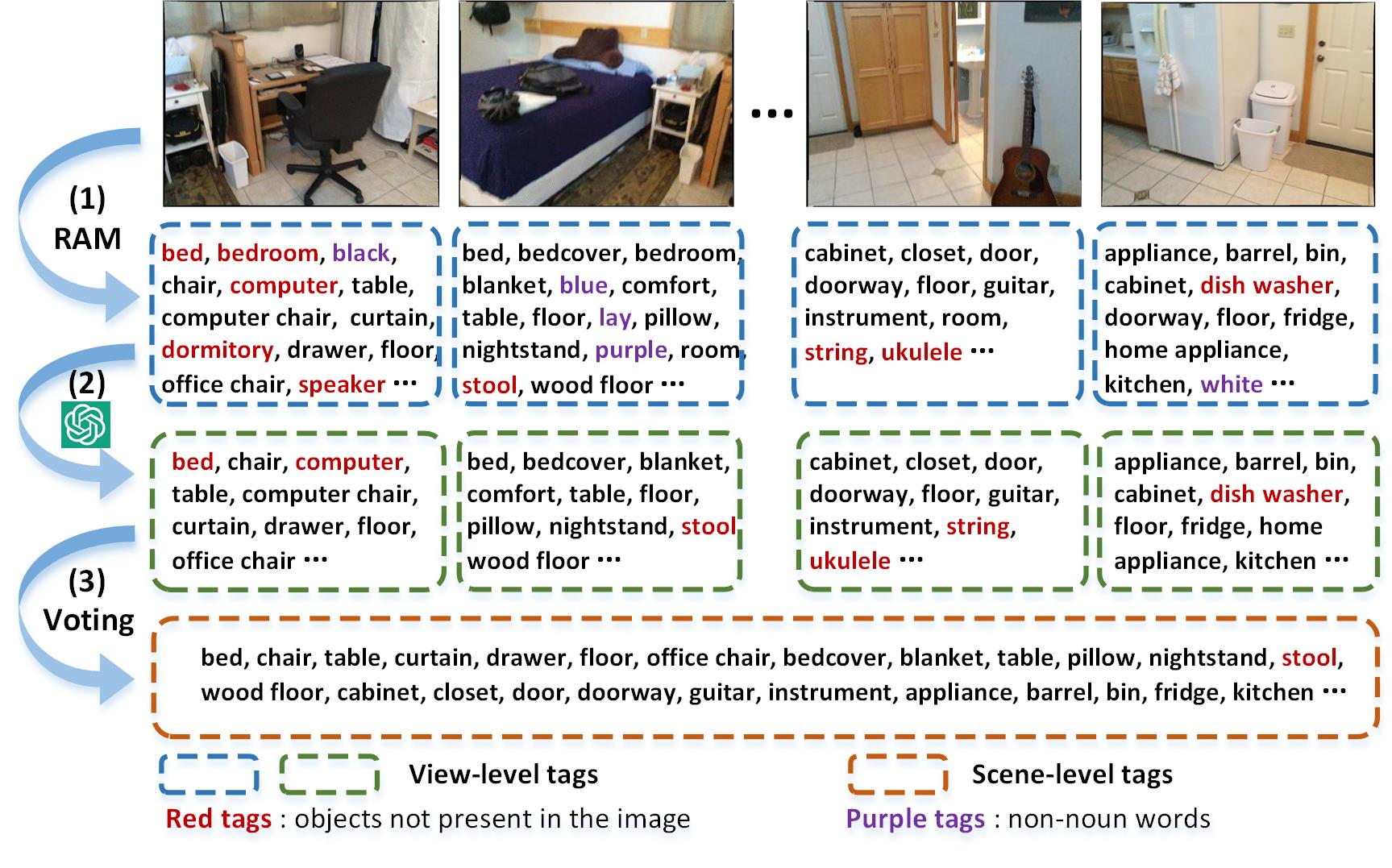}\\
	\vspace{-1em}
	\caption{Scene tagging generation. (1) We first employ RAM~\cite{zhang2023recognize} to generate view-level tags, and then (2) reduce the tag noise with GPT. Finally, scene-level tags are generated by (3) multi-view voting. }
	\vspace{-2em}
	\label{DMA:tags} 
\end{figure}

\textbf{Complete Category Information.} The scene tagging process is illustrated in Fig.~\ref{DMA:tags}. Firstly, we use the image tagging foundation model such as RAM~\cite{huang2023tag2text} to extract all possible categories from an image, and utilize category names and short descriptions derived from the metadata as the text query, referred to as $T_{tag}$, such as ``\texttt{There is a \{category name\} in the scene}'', ``\texttt{A photo of a \{category name\}}'', \etc. Unlike image captioning models~\cite{li2023blip,alayrac2022flamingo} that can only identify sparse and salient objects in a scene, RAM can recognize as many tags as possible without missing important parts, \textit{ensuring a high recall rate and alignment with complete semantic patterns}. The more complete and accurate the detected categories are, the easier we can establish precise dense correspondences between text and 3D modalities, and hence the open-vocabulary capability of the 3D model can be enhanced.% by aligning with a variety of categories.

\textbf{Reliable GPT-based Denoising.} While there are many tags recognized by RAM, some redundant or irrelevant tags are also included, such as non-noun words (``\textcolor[RGB]{114,51,158}{\textbf{purple}}'', ``\textcolor[RGB]{114,51,158}{\textbf{blue}}'', ``\textcolor[RGB]{114,51,158}{\textbf{lay}}'', \etc) and objects that do not exist in the image (``\textcolor[RGB]{171,4,15}{\textbf{bed}}'', ``\textcolor[RGB]{171,4,15}{\textbf{ukulele}}'', \etc), as shown in Fig.~\ref{DMA:tags}. We address this issue in two steps. Firstly, we utilize GPT to filter potential noisy vocabulary. Given the input list, we instruct GPT to examine the words one by one and perform reasoning according to the chain of thought, outputting a boolean list indicating whether a word is an outlier. Please refer to \textit{Fig.~1 of supplemental material} for the detailed instructions and examples to reduce the noisy tags. Secondly, to decrease the non-existent categories in a scene, we conduct multi-view voting and neglect categories that appear in fewer than five views. Please refer to \textit{Fig.~2 of the supplemental material} for the denoised scene tagging results and the corresponding visualizations of 3D label maps. 

\textbf{Scalable Scene Description.} Although scene-level tags have already covered most of the categories, the limited scalability and variation of category names hinders their provision of rich contexts and details. To address this limitation and enable arbitrary queries for 3D networks, we additionally leverage MLLMs such as LLaVA~\cite{liu2023llava} to generate diverse and linguistically expressible descriptions, denoted by $T_{llm}$. Owing to the exposure to a diverse range of linguistic patterns and contextual nuances, the MLLMs can generate comprehensive and in-depth descriptions based on input images. Consequently, these LLMs can enhance the richness and granularity of the generated textual representations, thereby facilitating a more comprehensive understanding of the 3D scenes. Please refer to \textit{Fig.~3 of supplemental material} for the examples of scene-level captions and corresponding visualizations.

Finally, we generate the text embeddings $\textbf{f}^T_{tag}$ and $\textbf{f}^T_{llm}$ using CLIP text encoder based on the generated tags $T_{tag}$ and scene descriptions $T_{llm}$, respectively, which are utilized to supervise the training of 3D networks subsequently.   
\begin{figure}[!t]
	\centering 
	\hspace{-1.0cm}
	\includegraphics[scale=0.3]{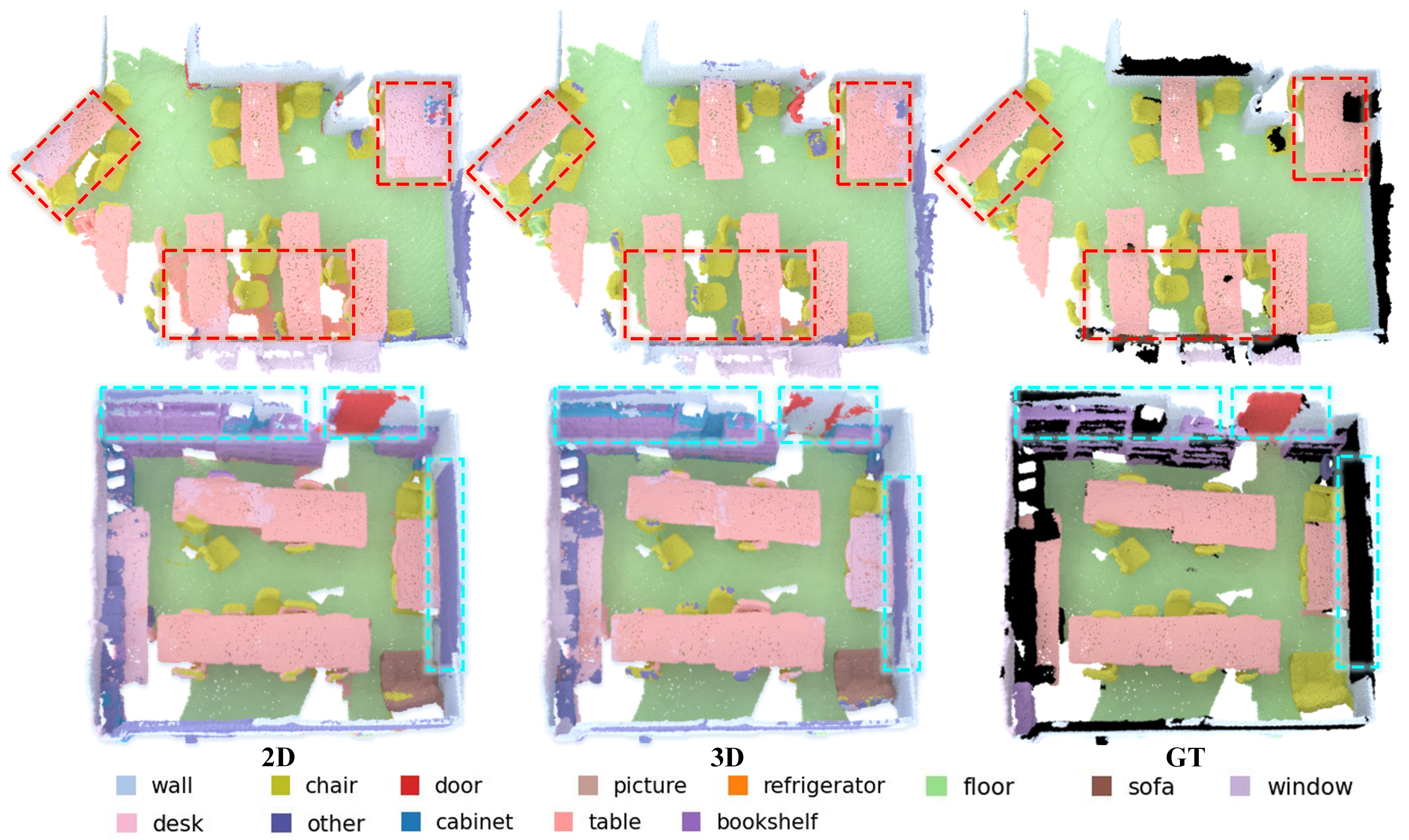}\\
	\caption{Segmentation results using 2D and 3D models. 2D model has advantages in segmenting background objects (in  \textcolor{cyan}{\textbf{blue}} boxes), while 3D model is more favorable for foreground objects with distinct structures (in  \textcolor{red}{\textbf{red}} boxes).}
	\vspace{-1em}
	\label{DMA:comparison} 
\end{figure}

\vspace{-0.2cm}
\subsection{Structure-aware Image Feature Extraction}     
Compared to language modality, the image modality offers a wealth of contextual information and exhibits significant variations among different pixels, which could provide more effective supervision. Inspired by this observation, OpenScene~\cite{peng2023openscene} distills the knowledge from frozen open-vocabulary 2D segmentation models, such as LSeg~\cite{li2022languagedriven} and OpenSeg~\cite{ghiasi2022scaling}. However, these methods suffer from two major limitations. Firstly, they are fine-tuned on in-vocabulary datasets, which leads to a misalignment between image and text features and consequently results in poor performance on open-vocabulary categories. Secondly, all of these methods freeze 2D networks, failing to perceive the 3D structure of objects and leading to inaccurate supervision. As shown in Fig.~\ref{DMA:comparison}, we visualize the segmentation results using 2D and 3D features. One can observe that although 2D features are more advantageous in segmenting background objects with ambiguous geometry, such as ``bookshelf'', ``door'' and ``blackboard'', they are less effective in segmenting objects with distinct shapes, such as ``table'' and ``chair''. Therefore, it is necessary to distill the structural priors of 3D networks into 2D ones as well in order to facilitate fine-grained scene understanding.
% but also limits the performance of 3D model. Indeed, as shown in Fig.~\ref{DMA:comparison}, 3D features have strong structural priors, which can be leveraged to improve the performance of 2D models in a reciprocal manner.  

%their mask prediction performance significantly lags behind query-based methods, such as ODISE~\cite{xu2023open} and FC-CLIP~\cite{yu2023fcclip}. 

In this paper, we adopt FC-CLIP~\cite{yu2023fcclip} as the backbone to extract image features. On one hand, we use the frozen CLIP visual encoder to ensure the intactness of image-text alignment, obtaining CLIP features $\textbf{f}^{2D}_{clip}$. On the other hand, to facilitate the synergistic benefits of both 2D and 3D modalities, we fine-tune the mask head and attain the mask features $\textbf{f}^{2D}_{mask}$. In contrast to previous methods that rely on potentially noisy fixed image features for supervision, the fine-tuned mask features enhance the adaptability to downstream 3D tasks. We explore different fine-tuning strategies, such as LoRA~\cite{hu2021lora}, Adapter~\cite{houlsby2019parameter}, and full parameter fine-tuning and compare them in experiments. 

%considering that CLIP models are typically pretrained on low-resolution images for classification tasks, which are difficult to be directly applied to downstream dense prediction tasks, 
%we additionally add the high-resolution mask features  to obtain the final image features, \ie, $\textbf{f}^{2D}=\textbf{f}^{2D}_{clip}+\textbf{f}^{2D}_{mask}$. Furthermore, different from previous methods utilizing fixed image features as supervision, we fine-tune the parameters of mask head to better adapt the segmentation model to different datasets. 
%In this way, image features can be improved by integrating the structural knowledge of 3D model, facilitating the synergistic benefits of both modalities. 

%given an RGB image with resolution $H\times W$,
\vspace{-0.2cm}
\subsection{Dense Associations across Modalities}
Once the text and image modalities are constructed, the subsequent step is to associate each point to its corresponding pixel and text. We utilize the image modality as a bridge to establish separate associations between pixels and other modalities. Firstly, we construct the associations between image and language modalities by taking $C$ different text embeddings $\textbf{f}^T=\{\textbf{f}^T_1,\cdots,\textbf{f}^T_C\}$ as classifier to assign text labels to each pixel, obtaining a 2D semantic score map, denoted by $S^{2D}\in \mathbb{R}^{H\times W\times C}$. This process can be formulated as follows:
\begin{align}
	S^{2D}_c(u,v) = \sigma(<\textbf{f}^{2D}(u,v), \textbf{f}^T_c>/\tau_1),
	\label{2D_probability}
\end{align} 
where $S^{2D}_c(u,v)$ denotes the probability that the pixel at location $(u,v)$ belongs to the $c$-th text label, and $<\cdot>$ represents the cosine similarity between two $\ell_2$-normalized feature vectors. $\tau_1$ is a temperature parameter.
Then we establish the associations between images and point clouds by back-propagating 3D points $\textbf{p}=(x,y,z)$ onto 2D positions $(u,v)$ using a projection matrix $T\in \mathbb{R}^{3\times 4}$, \ie, $[u,v,w]=T[x,y,z, 1]$, where $w$ is a scaling factor.

%We employ sigmoid $\sigma(\cdot)$ rather than Softmax operation to convert the feature distance to the semantic score. This is because Softmax assumes a mutually exclusive relationship between different text labels, while our generated text labels sometimes exhibit similar semantics (`suitcase' and `luggage') and inclusion relationships (`kitchen' and `refrigerator'), allowing for the possibility of one pixel belonging to multiple text labels simultaneously.        

%\noindent\textbf{Text-to-3D label assignment.} 
Finally, we associate the text and 3D modalities by taking image as the bridge. Given $K$ different projection views for one point $\textbf{p}$, we compute its average semantic score, denoted as $\bar{S}^{3D}$, across $K$ views $[S^{2D}_1(u_1,v_1), \cdots, S^{2D}_K(u_K,v_K)]$. Based on the aggregated 3D semantic scores, the final text-to-3D label map, denoted as $M^{3D}\in \mathbb{R}^{N\times C}$, can be derived by: 
\begin{align}
	M^{3D}_{c}=\left\{\begin{matrix}
		1, & \text{if} \ \  \bar{S}^{3D}_c>\text{threshold}\\
		0, &  else
	\end{matrix}\right.,
	\label{label_map}
\end{align}
where $N$ denotes the number of points and $M^{3D}_{c}$ indicates whether the point belongs to the $c$-th text label or not. $M^{3D}$ can be regarded as the pseudo label map for point cloud, serving as the supervision signal for training 3D models. 

It is noteworthy that instead of generating one-hot label through the \texttt{argmax} operation, we select all confident text labels whose scores exceed the threshold. This is because the generated text categories may exhibit similarities in semantics (like `suitcase' and `luggage') or inclusion relationships (such as `kitchen' and `stove'). As a result, it is highly possible that \textit{one single point corresponds to multiple text labels simultaneously}.  

\vspace{-0.2cm}
\subsection{Dense Multimodal Alignment}
\label{sec:alignment}
After obtaining the triplets of different modalities and their dense correspondences, the subsequent objective is to align the 3D points with their corresponding text and pixel embeddings. This alignment process involves several steps. Firstly, we extract 3D features for the point cloud by utilizing a 3D network, denoted as $\varepsilon_{3D}$. These features are then projected to match the dimension of the CLIP features. 

Next, we assign text labels to different 3D points by computing the cosine similarities between point and text embeddings $\textbf{f}^T$, yielding a 3D segmentation probability map $P^{3D}$:  
\begin{align}
	P^{3D}_{i,c} = \sigma(<\textbf{f}^{3D}_i, \textbf{f}^T_c>/\tau_2),
	\label{probability}
\end{align}  
where $\textbf{f}^{3D}_i$ denotes the feature of the $i$-th point, and $P^{3D}_{i,c}$ denotes the probability that the $i$-th point belongs to the $c$-th text label. Here we employ the Sigmoid activation function $\sigma(\cdot)$ since it will not lead to mutually exclusive relationships among different categories. 

{\bf Text-to-3D Supervision.} We use the text-to-3D label map $M^{3D}$ as the pseudo label to facilitate the alignment of point and text features. Different loss functions are employed for aligning the point embeddings with the tag ${\bf f}_{tag}^T$ and scene description ${\bf f}_{llm}^T$ embeddings. As can be seen in Fig.~\ref{DMA:framework}, we build dense associations between ${\bf f}_{tag}^T$ and the \textbf{entire point cloud}, resulting in $M_{tag}^{3D}$. Consequently, The Binary Cross Entropy (BCE) loss is used to effectively penalize both positive and negative samples:
\begin{align}
	\mathcal{L}_{3d-text (tag)} = \mathcal{L}_{BCE}(P^{3D},M_{tag}^{3D}).
\end{align}
As for ${\bf f}_{llm}^T$, since it corresponds only to salient objects, we can only obtain the mask for \textbf{partial points}, denoted as $M_{llm}^{3D}$. (Visualizations of $M_{tag}^{3D}$ and $M_{llm}^{3D}$ are given in \textit{Fig.~2 and Fig.~3 of the supplementary material}, respectively.) We utilize the cosine similarity loss to supervise only the positive samples:
\begin{align}
	\mathcal{L}_{3d-text (llm)} = 1-\cos({\bf f}_{llm}^T, \textbf{f}^{3D}).
\end{align}
{\bf Mutually Inclusive Loss.} In this work, we do not employ the Cross-Entropy loss because it would result in a mutually exclusive relationship between different classes, meaning that each point is assigned to only one class of interest. However, in text-to-3D alignment, one point may simultaneously associate with multiple text prompts, such as `bed' and `bedroom', `chair' and `office chair', `curtain' and `drape', \etc. To handle this issue, we employ mutually inclusive losses (\textbf{MIL}), such as BCE loss and cosine similarity loss, to ensure that each point is aligned with all its corresponding tags/descriptions simultaneously, avoiding the potential conflicts between categories with overlapping or similar semantics.

{\bf 2D-to-3D Supervision.} For 3D-2D pairs, we follow the previous work~\cite{peng2023openscene} to fuse pixel embeddings across $K$ different views, represented as $[\textbf{f}^{2D}_1,\cdots, \textbf{f}^{2D}_K]$, into a single feature vector $\bar{\textbf{f}}^{2D}$, and align 2D and 3D features by minimizing the cosine similarity loss:
\begin{align}
	\mathcal{L}_{3d-2d} = 1-\cos(\bar{\textbf{f}}^{2D}, \textbf{f}^{3D}).
\end{align}  

Since 2D mask head is also trainable, we additionally add the text-to-2D supervision and compute the BCE loss between 2D predictions and 2D masks, obtaining $\mathcal{L}_{text-2d}$. Finally, the overall objective function to perform dense multimodal alignment is defined as:
\begin{align}
	\mathcal{L}_{3D} = \mathcal{L}_{3d-text(tag)} + \mathcal{L}_{3d-text(llm)} + \mathcal{L}_{3d-2d} + \mathcal{L}_{text-2d},
\end{align}
where the language modality provides comprehensive textual descriptions, and the image modality gives precise supervision on object edges and contextual information. Additionally, the 3D modality reveals crucial structural information of objects. By densely aligning these modalities in a shared space, our method can maximize the synergistic benefits among them and achieve outstanding segmentation performance without compromising the open-vocabulary classification ability of the model.  

\section{Experiments}

\subsection{Setups} 
\textbf{Datasets.} To demonstrate the effectiveness of our proposed method, we employ three popular datasets,~\ie, ScanNet~\cite{dai2017scannet}, Matterport3D~\cite{chang2017matterport3d}, and nuScenes~\cite{caesar2020nuscenes}. The first two datasets are indoor ones, comprising RGBD images and 3D meshes. The third one is an outdoor dataset, consisting of data collected from two sensors,~\ie, LiDAR and camera. We conduct comparisons with state-of-the-art methods on each of these datasets. The mean Intersection-of-Union (mIoU), mean Accuracy (mACC), Precision, and Recall are employed as the evaluation metrics. 

\textbf{Implementation Details.} In this work, MinkowskiNet~\cite{choy20194d} is employed as the 3D backbone, whose voxel size is set to 2cm for ScanNet and Matterport3D and 5cm for nuScenes. As for the 2D backbone, we use OpenSeg~\cite{ghiasi2022scaling} and FC-CLIP~\cite{yu2023fcclip} that perform mask-wise classification. The parameters $\tau_1$ in Eq.~\ref{2D_probability} and $\tau_2$ in Eq.~\ref{probability} are both set to 0.1. We use Adam~\cite{kingma2014adam} as the optimizer and the initial learning rate is set to $1e-4$. The model is trained for 100 epochs. We set the batch size as 8 for indoor datasets and use one single NVIDIA RTX A6000 for training. As for nuScenes dataset, we use 8 GPUs for training and set the batch size as 16. 
\begin{table}[!t]
	\hspace{-0.5cm}
	\begin{minipage}[htbp]{\textwidth}
		\begin{minipage}[t]{0.5\textwidth}
			\centering
			\makeatletter\def\@captype{table}\makeatother  
			\scalebox{0.66}{
				\begin{tabular}{p{0.3cm}|c|cc|cc|c}
					\toprule \rowcolor{cyan!15}
					& {\scriptsize Methods}                     & {\scriptsize mIoU} & {\scriptsize mACC} & {\scriptsize mIoU{\tiny(F)}} & {\scriptsize mIoU{\tiny (B)}}  & {\scriptsize Latency} \\ \hline \hline
					\multirow{6}{*}{\textit{\rotatebox{90}{\scriptsize fully-supervised}}}
					&TangentConv~\cite{tatarchenko2018tangent}                & 40.9 &    $-$  &    $-$        &   $-$   &   $-$      \\
					&TextureNet~\cite{huang2019texturenet}                   & 54.8 &  $-$   &   $-$         &  $-$   &   $-$      \\
					&ScanComplete~\cite{dai2018scancomplete}                 & 56.6 &  $-$    &   $-$         &  $-$  &   $-$        \\
					&Mix3D~\cite{nekrasov2021mix3d}                        & \textbf{73.6} &  $-$    &     $-$       &    $-$   &   $-$     \\
					&VMNet~\cite{hu2021vmnet}                        & 73.2 &  $-$    &    $-$        &   $-$   &   $-$      \\
					&MinkowskiNet~\cite{hu2021vmnet}                 & 69.0 &  $-$    &   $-$         &  $-$   &   $-$       \\ \hline
					\multirow{7}{*}{\textit{\rotatebox{90}{\scriptsize Zero-shot }}}
					&PLA~\cite{ding2023pla}                          & 17.7 & 33.5 &     $-$       &   $-$    &    0.07s     \\
					&RegionPLC~\cite{yang2023regionplc}                    & 43.8 & 65.6 &    $-$        &   $-$   & 0.07s       \\ \cdashline{2-7}
					&{\scriptsize OpenScene~\cite{peng2023openscene}{\tiny (LSeg)}-3D}		 & 52.9 & 63.2 &     $-$       &   $-$  & 0.07s     \\
					&{\scriptsize OpenScene{\tiny (LSeg)}-2D3D}		 & {54.2} &  66.6 &     $-$       &   $-$  &  102.6s      \\ 
					&{\scriptsize OpenScene$^\dag${\tiny (OpenSeg)}-3D}				 & 46.6 & 66.5 &     50.0       &   47.1   &  0.07s     \\
					&{\scriptsize OpenScene$^\dag${\tiny (OpenSeg)}-2D3D}				 & 47.9 & \textbf{71.7} &     49.5       &   51.0    &  89.4s     \\
					\cdashline{2-7}
					&{\scriptsize DMA{\tiny (OpenSeg)}-text only}     				 & 50.5 & 63.7 &     56.7      &    48.0  & 0.07s      \\
					&{\scriptsize DMA{\tiny (OpenSeg)}-3D}     				 & 53.3 & 70.3 &     {58.3}      &    {51.5}   & 0.07s      \\
					&{\scriptsize DMA{\tiny (LSeg)}-3D}     				 & \textbf{54.8} & 66.9 &     \textbf{59.9}      &    \textbf{51.9}   & 0.07s      \\
					&{\scriptsize DMA{\tiny (FC-CLIP)}-3D}     				 & 51.8 & 68.7 &     56.0       &   51.4   & 0.07s    \\ \bottomrule
			\end{tabular}}
			\caption{Comparison on the ScanNet~\cite{dai2017scannet} validation set. ``F'' and ``B'' denote foreground and background classes, respectively. $^\dag$ denotes our reproduced results.}
			\label{tab:scannet}
		\end{minipage}
		\hspace{1.2em}
		\begin{minipage}[t]{0.5\textwidth}
			\centering
			\makeatletter\def\@captype{table}\makeatother
			\scalebox{0.79}{
				\begin{tabular}{c|c|p{0.7cm}|c|p{0.8cm}p{1.3cm}}
					\toprule \rowcolor{cyan!15}
					& \multirow{2}{*}{{\scriptsize Methods}}           & \multirow{2}{*}{{\scriptsize Anno.}}                & \multirow{2}{*}{\scriptsize mIoU} & {\scriptsize mIoU {\tiny (Base)}} & {\scriptsize mIoU {\tiny(Long-Tail)}} \\ \hline\hline
					\multirow{4}{*}{\textit{\rotatebox{90}{\scriptsize fully-sup.}}}
					& RangeNet++~\cite{milioto2019rangenet++}        & \multirow{4}{*}{100\%} & 65.5 &       76.4      &    56.4              \\
					& Cylinder3D~\cite{zhu2021cylindrical}        &                      & 75.4 &    \textbf{ 84.1}        &  69.4                \\
					& SPVNAS~\cite{tang2020searching}          &                      & 74.8     &     82.3        &      67.2            \\
					& AMVNet~\cite{liong2020amvnet}            &                      & \textbf{77.0} &      83.9       &  \textbf{70.8}                \\ \hline
					\multirow{4}{*}{\textit{\rotatebox{90}{\scriptsize weakly-sup.}}}
					& ContrastiveSC~\cite{hou2021exploring}     & \multirow{2}{*}{0.9\%} & 64.5     &    79.7         &  53.8                \\
					& LESS~\cite{liu2022less}              &                      & \textbf{74.8}   &    \textbf{81.6}         &     \textbf{68.7 }            \\ \cdashline{2-6}
					& ContrastiveSC     & \multirow{2}{*}{0.2\%} &  63.5    &  78.4           &  51.6                \\
					& LESS              &                      &  73.5    &  81.1           &    66.6              \\ \hline
					\multirow{4}{*}{\textit{\rotatebox{90}{\scriptsize Zero-shot}}}
					& {\scriptsize OpenScene~\cite{peng2023openscene}{\tiny (LSeg)}-2D3D}    & \centering\multirow{2}{*}{No} &  36.7 &   55.0      &  22.3                \\
					& {\scriptsize OpenScene{\tiny (OpenSeg)}-2D3D} &                     &  42.1 & 52.6            &    33.8             \\ \cdashline{2-6}
					& DMA{\tiny (OpenSeg)}-3D   & \centering\multirow{2}{*}{No}  &  45.1 &   59.3          &   33.9               \\
					& DMA{\tiny (FC-CLIP)}-3D       &                     &  \textbf{47.4} &    \textbf{61.4}         &    \textbf{35.3}              \\ \bottomrule
			\end{tabular}}
			\caption{Comparison on the nuScenes~\cite{caesar2020nuscenes} validation set. We partition all categories into base and long-tail classes according to their frequencies.  }
			\label{tab:nuscenes}
		\end{minipage}
	\end{minipage}
	\vspace{-1.0cm}
\end{table}

\subsection{Comparison with State-of-the-Arts}
We compare the proposed DMA with fully-/weakly-supervised and zero-shot methods~\cite{peng2023openscene,ding2023pla,yang2023regionplc}. Tab.~\ref{tab:scannet} presents the segmentation results on the \textit{\textbf{ScanNet}}~\cite{dai2017scannet} dataset. To facilitate comparison, we measure the results of OpenScene by using 3D and 2D-3D integrated features as supervision. As can be seen in Tab.~\ref{tab:scannet}, although OpenScene(LSeg) attains better results (54.2\% mIoU) by using both 2D and 3D encoders, it results in significantly \textbf{increased inference latency}. This is because the parameter size of 2D encoder is much larger than 3D encoder, and the 2D encoder needs to perform inference on multi-view images of the scene. Our DMA(OpenSeg) using only 3D model for prediction outperforms OpenScene(OpenSeg)-2D3D by 5.4\% mIoU at a significantly lower latency, wherein the mIoU (F) and mIoU (B) are improved by 8.8\% and 0.5\%, respectively. This is because we perform additional alignment with text modality, thereby compensating for the decreased open-vocabulary ability of 2D model. When using text supervision only, our method outperforms the text-supervised approach RegionPLC~\cite{yang2023regionplc} by 9.5\%, and even surpasses OpenScene(OpenSeg)-2D3D by 2.6\% in terms of mIoU. This indicates that, compared to previous methods that generate image- and region-level captions, our method establishes dense and precise correspondences between text and 3D points by taking 2D modality as the bridge, achieving more precise supervision. The suboptimal performance of our method using FC-CLIP as the 2D encoder may be attributed to the low resolution of the images (320$\times$240), which limits the capabilities of FC-CLIP.

\begin{figure*}[!t]
	\centering 
	\hspace{-0.8cm}
	\includegraphics[scale=0.44]{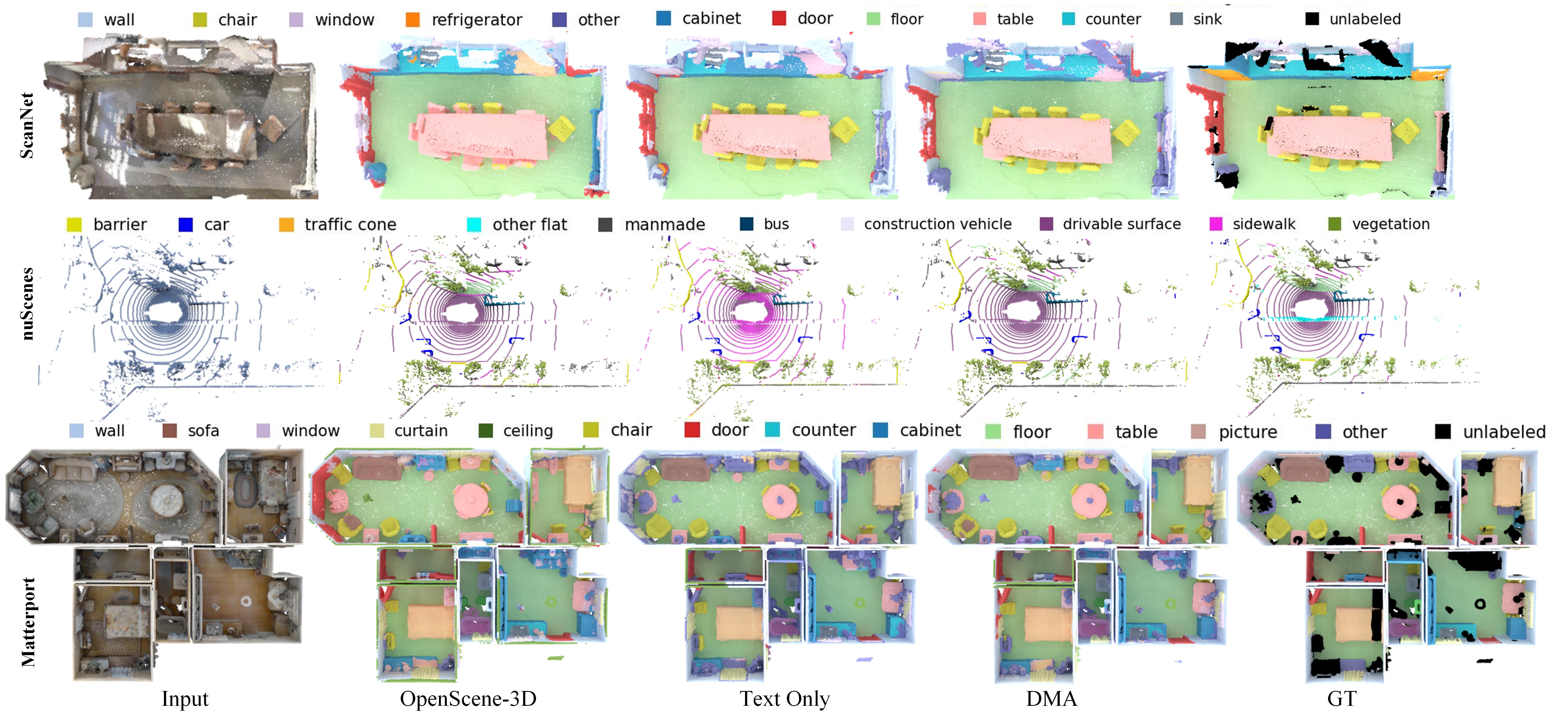}\\
	\vspace{-1em}
	\caption{Qualitative results of different methods on both indoor and outdoor datasets.}
	\label{DMA:qualitative} 
	\vspace{-0.2cm}
\end{figure*}

\textbf{Outdoor Scenes.} To validate the effectiveness of our method on outdoor point clouds, we evaluate the performance of DMA on the \textbf{\textit{nuScenes}} dataset~\cite{caesar2020nuscenes}. Due to the highly imbalanced class distribution of outdoor scenes, we additionally measure the performance on base and long-tail categories. As shown in Tab.~\ref{tab:nuscenes}, by densely aligning with the tagging information and the detailed description extracted from each scene, our DMA(OpenSeg) using only 3D encoder significantly improves the performance over OpenScene(OpenSeg)-2D3D by 3.0\% mIoU. Additionally, the final performance is further improved by 2.3\% and attains 47.4\% mIoU by employing FC-CLIP~\cite{yu2023fcclip} to extract 2D features. This is because FC-CLIP could achieve more precise segmentation while maintaining outstanding open-vocabulary recognition ability of CLIP. Besides, by fine-tuning the mask head, FC-CLIP could incorporate the 3D structural priors into mask features and produce better results.

\begin{table*}[!t]
	\centering
	\scalebox{0.73}{
		\begin{tabular}{c|cccc|cccc|cccc|cccc}
			\toprule  \rowcolor{cyan!15}
			& \multicolumn{4}{c}{mIoU}    & \multicolumn{4}{c}{mACC} & \multicolumn{4}{c}{Precision} & \multicolumn{4}{c}{Recall}   \\ \hline
			& Head & Common & Tail & All  & Head & Common & Tail & All & Head & Common & Tail & All & Head & Common & Tail & All \\ \hline \hline
			{\scriptsize OpenScene}{\tiny$^\dag$\cite{peng2023openscene}-2D3D}   &  21.2  &   8.4     &   4.0   &   6.2   &  35.5    & 15.7   &   9.4   & 12.0  & 43.9  & 18.6     &  7.8 &  14.5  &    34.5     & 15.7 & 9.4  &12.0         \\
			PLA~\cite{ding2023pla}         & $-$  &  $-$   & $-$  &  1.8 &  $-$ & $-$    &  $-$ & 3.1  &  $-$ & $-$    &  $-$ & $-$ &  $-$ & $-$    &  $-$ &  $-$  \\
			RegionPLC~\cite{yang2023regionplc}   & $-$  &  $-$   & $-$  &  6.5 &  $-$ & $-$    &  $-$ & 15.9  &  $-$ & $-$ &  $-$ & $-$    &  $-$ &  $-$  &  $-$ &  $-$ \\ \cdashline{1-17}
			DMA-text only&	23.2   &  7.6	&	2.0   & 6.9	  &	 32.6 &	13.4 &	5.9	 & 11.3 & 40.0 &  15.1 & 5.8  & 13.7 & 32.6 & 13.4 & 5.9 & 11.3\\
			{\scriptsize DMA(OpenSeg)-3D} &  25.3  &  10.8  &  5.5  &  7.6  &  36.7 &   18.2 &  10.7  &   14.6  &  44.5 &  23.6  &  10.5  &   14.9 &   36.7 &  18.2    &  10.7  &   13.2 \\
			{\scriptsize DMA(FC-CLIP)-3D} &  27.2  &  11.5   &  5.8   &  7.9    &  37.4    &  19.2  & 11.2    & 15.2 & 46.2 & 24.9 & 11.3  & 15.7  & 38.2 & 20.4 & 11.1  & 14.0  \\ \hline
			Fully-Sup   & 45.5 &  13.6  & 3.4  & 20.8 & $-$  &  $-$    & $-$  & $-$ & 66.8 & 55.7 & 23.3 & 34.4 & 57.6 & 19.1 & 5.8 & 27.8  \\ \bottomrule
	\end{tabular}}
	\caption{Comparison on ScanNet200~\cite{rozenberszki2022language} validation set. $\dag$ means our reproduced results.}
	\label{tab:scannet200}
	\vspace{-0.6cm}
\end{table*}

\textbf{Long-Tail Datasets.} As shown in Tab.~\ref{tab:scannet200}, we validate the open-vocabulary methods on the more challenging long-tail 3D scene understanding datasets, \ie, \textbf{\textit{ScanNet200}}~\cite{rozenberszki2022language}. Following~\cite{rozenberszki2022language}, we partition the 200 categories into three splits, \ie, \texttt{head}, \texttt{common}, and \texttt{tail} sets, facilitating a more comprehensive comparison across categories with different frequencies. When training on \texttt{head} classes (ceiling, curtain, window, \etc), the fully-supervised method performs much better than zero-shot methods due to the sufficient 3D labels for supervision. However, on the \texttt{common} and \texttt{tail} splits, our DMA method approaches to or even surpasses the fully-supervised competitors. This is because there are only a few instances available on these long-tail categories, which is not sufficient to train a robust model from scratch. Our method does not rely on ground truth 3D labels but instead distill knowledge from pretrained vision-language models, thus it is more robust to rare objects. 

To further validate the robustness of our method on rare objects/classes, we evaluate on the most frequent $K$ classes of \textbf{\textit{Matterport3D}}~\cite{chang2017matterport3d}, where $K=21, 40, 80, 160$. We train a 3D model by taking our generated textual descriptions and image features as supervision, and perform inference on different $K$ categories. As shown in Tab.~\ref{tab:matterport}, when employing the same 2D network, \ie, OpenSeg, our method demonstrates superior zero-shot segmentation capability on both common and rare categories. Specifically, our DMA(OpenSeg)-3D surpasses OpenScene(OpenSeg)-3D by 3.8\%, 4.5\%, 1.6\%, and 1.2\% in terms of mIoU at different $K$. This can be attributed to that OpenScene heavily relies on 2D model for supervision without aligning with text prompts, which limits its open-vocabulary ability. Our method, however, directly aligns with the textual modality, overcoming the limitations of 2D models. 

\begin{table}[!t]
	\centering
	\scalebox{0.8}{
		\begin{tabular}{c|c|cccc|cccc}
			\toprule  \rowcolor{cyan!15}
			&  &  \multicolumn{4}{c}{mIoU}    & \multicolumn{4}{c}{mACC}   \\ \hline
			& \# of classes K & 21 & 40 & 80 & 160  & 21 & 40 & 80 & 160 \\ \hline \hline
			\multirow{6}{*}{\textit{\rotatebox{90}{\small fully-sup.}}}
			& TangentConv~\cite{tatarchenko2018tangent}& $-$ & $-$ & $-$ & $-$ & 46.8 &  $-$ & $-$ & $-$                         \\
			& TextureNet~\cite{huang2019texturenet}    &   $-$ & $-$ & $-$ &  $-$ & 63.0 &  $-$ & $-$ & $-$                          \\
			& ScanComplete~\cite{dai2018scancomplete}  &   $-$ & $-$& $-$&  $-$  & 44.9     &  $-$ & $-$ & $-$                   \\
			& DCM-Net           &  $-$  & $-$& $-$&  $-$   & 66.2 &    $-$ & $-$ & $-$                  \\
			& VM-Net~\cite{hu2021vmnet}           &   $-$ & $-$& $-$&  $-$ & \textbf{67.2} &  $-$ & $-$ & $-$                  \\
			& MinkowskiNet~\cite{hu2021vmnet} &     54.2 & $-$& $-$& $-$  & 64.6 &  $-$ & $-$ & $-$                  \\ \hline
			\multirow{6}{*}{\textit{\rotatebox{90}{\small Zero-shot.}}}
			& OpenScene~\cite{peng2023openscene}(LSeg)-3D   & 41.9 &   25.4   & 12.0  &  5.9  & 51.2 & 30.7  &  15.2& 7.5 \\
			& OpenScene(LSeg)-2D3D   & 43.4 &  26.8  & 13.1  &  6.4  & 53.5 &  33.0  & 17.4 & 8.6 \\
			& OpenScene(OpenSeg)-3D   & 41.3 & 33.4  &  18.1 &  8.2    &55.1  &  46.7      &  26.2 & 13.9    \\
			& OpenScene(OpenSeg)-2D3D   & 42.6 & 34.2  & 18.8  & 8.4     &\textbf{59.2}  &   47.5    & \textbf{27.1}  & 14.5    \\
			\cdashline{2-10}
			& DMA-text only& 39.8 &	25.4 & 11.7  & 6.2  &	49.5 &	31.6  &	16.1 & 8.0   \\
			& DMA(OpenSeg)-3D &  45.1  &  37.9   &  19.7    &   9.4   &  57.6    &  47.7  & {26.7}  & 14.1    \\
			& DMA(FC-CLIP)-3D &   \textbf{46.2}   &    \textbf{38.4}    &  \textbf{20.1}   &  \textbf{9.8}   &   {58.4}   &   \textbf{48.3}     &  26.5    &  \textbf{15.2 }  \\ \bottomrule
	\end{tabular}}
	\caption{Comparison on the Matterport~\cite{chang2017matterport3d} test set. }
	\label{tab:matterport}
	\vspace{-0.6cm}
\end{table}

\textbf{Qualitative Comparison.} Fig.~\ref{DMA:qualitative} visualizes the segmentation results of different methods. We can observe that OpenScene~\cite{peng2023openscene} with only 3D encoder exhibits poor performance in segmenting objects that lack spatial structures, such as ``door'', ``window'', ``counter'', \etc. In contrast, text supervision offers more refined guidance by establishing dense correspondences between the texts and points, thereby enabling more precise alignment. Our approach leverages the advantages of both language and 2D modalities, and achieves excellent segmentation results for both foreground and background classes using only the 3D model.

%OpenScene-2D denotes that only the multi-view image features are fused for prediction, which will lead to heavy inference overhead.      

%\begin{table}[!t]
%	\centering
%	\scalebox{0.8}{
	%	\begin{tabular}{c|ccc}
		%		\toprule \rowcolor{cyan!15}
		%		& ScanNet & Matterport3D & nuScenes \\ \hline \hline 
		%		Tagging & 46.3    &              & 43.3     \\
		%		Caption & 24.2    &              & 11.3     \\
		%		Both    & 50.5    &              & 43.8    \\ \bottomrule
		%	\end{tabular}}
%\caption{Comparison of different text descriptions.}
%\label{tab:tagging}
%\end{table}

\subsection{Ablation Study}
\begin{figure}[!t]
	\hspace{-0.5cm}
	\begin{minipage}[htbp]{\textwidth}
		\begin{minipage}[t]{0.65\textwidth}
			\centering
			%			\makeatletter\def\@captype{table}\makeatother  
			\includegraphics[scale=0.38]{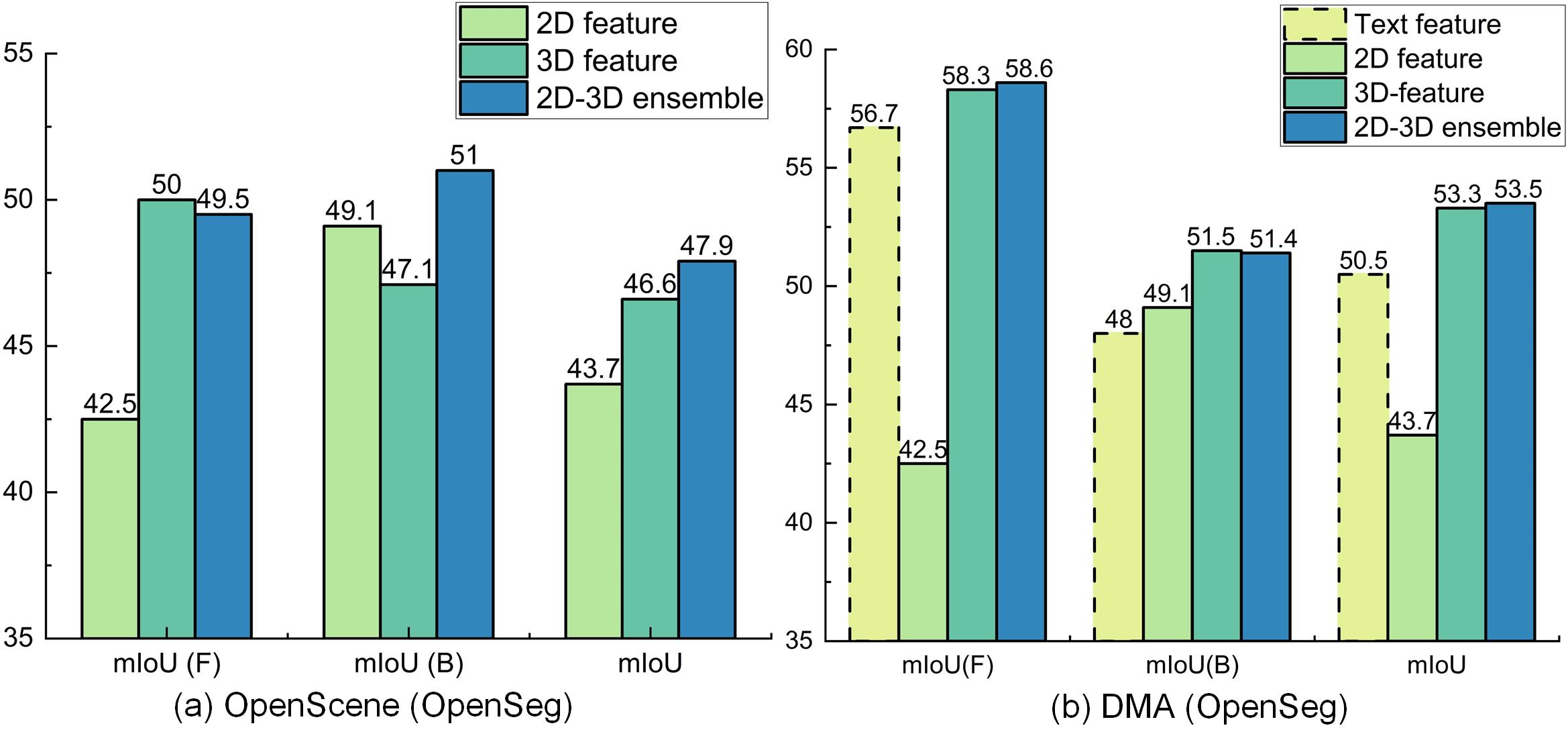}
			\vspace{-0.3cm}
			\caption{Comparisons of text, 2D, and 3D features. ``F'' and ``B'' denote foreground and background classes.}
			\label{DMA:2d_feature}
		\end{minipage}
		\hspace{0.3em}
		\begin{minipage}[t]{0.35\textwidth}
			\centering
			\vspace{-3.65cm}
			%			\makeatletter\def\@captype{table}\makeatother
			\includegraphics[scale=0.206]{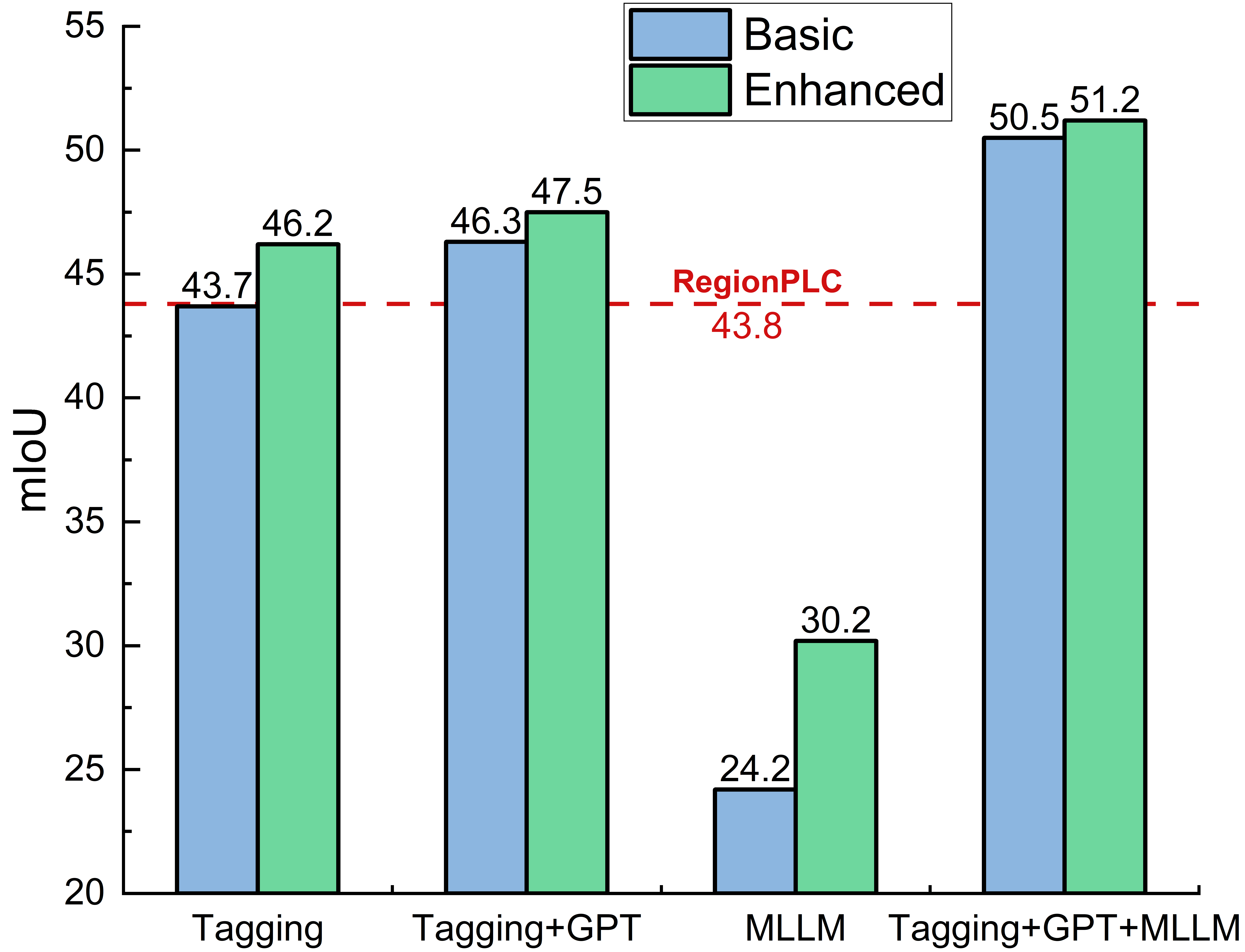}
			\vspace{-1.0mm}\vspace{-0.3cm}
			\caption{Comparisons of tagging models and MLLMs.}
			\label{DMA:VLM}
		\end{minipage}
	\end{minipage}
	\vspace{-0.3cm}
\end{figure}

{\bf 2D Features \vs 3D Features.} In Fig~\ref{DMA:2d_feature}, we compare the segmentation performance on ScanNet by using different features. `F' and `B' denote foreground and background classes, respectively. For OpenScene~\cite{peng2023openscene}, we observe that its 2D features are more advantageous for segmenting background categories with ambiguous geometry than 3D ones, \ie, 49.1\% \vs 47.1\% mIoU(B), while 3D features excel at segmenting the foreground objects with distinct shapes, \ie, 50.0\% \vs 42.5\% mIoU(F). Although the 2D-3D hybrid feature can leverage the strengths of both features simultaneously, utilizing 2D models for inference introduces significant computational overhead (please refer to the latency in Tab.~\ref{tab:scannet}). By additionally aligning with our generated text modality, our method can achieve outstanding performance on both foreground (58.3\%) and background (51.5\%) categories using only 3D features. Besides, DMA achieves comparable performance to using both 2D and 3D encoders by solely utilizing the 3D encoder, \ie, 53.3\% \vs 53.5\% mIoU(F), and hence significantly reducing inference time. 

{\bf Tagging Models \vs MLLMs.} In Tab.~\ref{DMA:VLM}, we compare the results of using different tagging models and MLLMs on ScanNet. For the enhanced version, we replace RAM with RAM++~\cite{huang2023open}, and LLaVA-7B with LLaVA-13B. We can observe that our method outperforms RegionPLC~\cite{yang2023regionplc} by a large margin (about 6.7\%) by building dense point-to-text correspondences. The tagging model plays a key role for performance improvement since it encompasses extensive semantic patterns, while MLLM further enhances the final performance by incorporating rich contextual information. By filtering out noisy tags with GPT, the performance can be improved by 2.6\% and 1.3\% for the basic and the enhanced versions, respectively. The final performance can be further improved when stronger tagging models/MLLMs are employed. 
%This is because we additionally align 3D representations with scalable and abundant text features, enhancing the comprehensive understanding of 3D scenes.   

%\begin{figure}[!t]
%	\centering 
%	\includegraphics[scale=0.44]{2d_feature_2.jpg}\\
%	\caption{Comparison of different features on ScanNet dataset. ``Text feature'' denotes the performance of 3D model trained only with text supervision. Both approaches (OpenScene and DMA) employ the same 2D backbone OpenSeg~\cite{ghiasi2022scaling}.}
%	\label{DMA:2d_feature} 
%\end{figure}

\begin{table}[!t]
	\hspace{-0.5cm}
	\begin{minipage}[htbp]{\textwidth}
		\begin{minipage}[t]{0.65\textwidth}
			\centering
			\makeatletter\def\@captype{table}\makeatother  
			\scalebox{0.8}{
				\begin{tabular}{c|cc|cc}
					\toprule \rowcolor{cyan!15}
					Method                & mIoU & mACC & mIoU (In) & mIoU (Out) \\ \hline \hline
					CLIP feature          & 35.2 & 51.3 & 36.7      & 31.8       \\
					Mask feature (w/o FT) & 40.1 & 55.4 & 48.3      & 21.3       \\
					Mask feature (w/ FT)  & 42.0 & 57.4 & 50.5      & 24.1       \\
					CLIP+Mask             & 44.8 & 59.7 & 51.7      & 28.5     \\ \bottomrule
			\end{tabular}}
			\caption{Comparisons of CLIP and Mask features of FC-CLIP on ScanNet. ``FT'' denotes fine-tuning.} 
			\label{DMA:fcclip_feature}
		\end{minipage}
		\hspace{0.3em}
		\begin{minipage}[t]{0.35\textwidth}
			\centering
			\makeatletter\def\@captype{table}\makeatother
			\scalebox{0.8}{
				\begin{tabular}{c|cc}
					\toprule \rowcolor{cyan!15}
					Method                & {\scriptsize 2D~Mask} & {\scriptsize 3D~Mask}  \\ \hline \hline
					w/o FT          & 36.6 & 40.1       \\
					Full Parameter & 40.4 & 42.0       \\
					LoRA~\cite{hu2021lora}  & 39.0 & 41.3       \\
					Adapter~\cite{houlsby2019parameter}  & 37.9 & 40.9       \\ \bottomrule
			\end{tabular}}
			\caption{Comparisons of different fine-tuning methods.}
			\label{DMA:finetune}
		\end{minipage}
	\end{minipage}
	\vspace{-1.0cm}
\end{table}

%\begin{table}[!t]
%	\centering
%	\scalebox{0.8}{
	%		\begin{tabular}{c|cc|cc}
		%			\toprule \rowcolor{cyan!15}
		%			Method                & mIoU & mACC & mIoU (In) & mIoU (Out) \\ \hline \hline
		%			CLIP feature          & 35.2 & 51.3 & 36.7      & 31.8       \\
		%			Mask feature (w/o FT) & 40.1 & 55.4 & 48.3      & 21.3       \\
		%			Mask feature (w/ FT)  & 42.0 & 57.4 & 50.5      & 24.1       \\
		%			CLIP+Mask             & 44.8 & 59.7 & 51.7      & 28.5     \\ \bottomrule
		%	\end{tabular}}
%	\caption{Comparison of different features of FC-CLIP on ScanNet. ``In'' and ``Out'' denote in-vocabulary and out-vocabulary, respectively. ``FT'' denotes fine-tuning.} 
%	\label{DMA:fcclip_feature}
%	\vspace{-0.8cm}
%\end{table}
%
%\begin{table}[!t]
%	\centering
%	\scalebox{0.8}{
	%		\begin{tabular}{c|cc}
		%			\toprule \rowcolor{cyan!15}
		%			Method                & {\scriptsize 2D~Mask} & {\scriptsize 3D~Mask}  \\ \hline \hline
		%			w/o FT          & 44.6 & 40.1       \\
		%			Full Parameter & 48.9 & 42.0       \\
		%			LoRA  & 47.0 & 41.3       \\
		%			Adapter  & 46.3 & 40.9       \\ \bottomrule
		%	\end{tabular}}
%	\caption{Comparison of different features of FC-CLIP on ScanNet. ``In'' and ``Out'' denote in-vocabulary and out-vocabulary, respectively. ``FT'' denotes fine-tuning.} 
%	\label{DMA:finetune}
%	\vspace{-0.8cm}
%\end{table}

{\bf CLIP Features \vs Mask Features. } In addition to OpenSeg~\cite{ghiasi2022scaling}, we employ FC-CLIP~\cite{yu2023fcclip} to extract 2D features due to its effectiveness. As show in Tab.~\ref{DMA:fcclip_feature}, we compare the performance by using CLIP and Mask features as supervision. `In' and `Out' denote in-vocabulary and out-vocabulary classes, respectively. FC-CLIP contains an in-vocabulary classifier and an out-vocabulary classifier, which correspond to the seen and unseen categories in the training process, respectively. As can be seen, the fixed CLIP feature is more advantageous in segmenting unseen categories, which outperforms mask feature by 10.5\% in terms of mIoU(Out). This demonstrates that the fixed CLIP visual encoder could maintain the strong generalization ability on novel classes. While for in-vocabulary classes, mask features outperform the CLIP feature by 11.6\%. By combining these features, we can simultaneously achieve competitive results on in- and out-vocabulary categories, attaining 44.8\% mIoU over all classes. 

{\bf Comparisons of Different Fine-Tuning Methods.} We fine-tune the mask head of FC-CLIP with different strategies for incorporating 3D structural priors into mask features. As can be seen in Tab.~\ref{DMA:finetune}, by fully fine-tuning the mask head, the performances of 2D and 3D masks are improved by 3.8\% and 1.9\%, respectively. LoRA~\cite{hu2021lora} and Adapter~\cite{houlsby2019parameter} can also achieve obvious improvements by tuning a small amount of parameters. 

\begin{figure}[!t]
	\centering
	\includegraphics[width=0.75\textwidth]{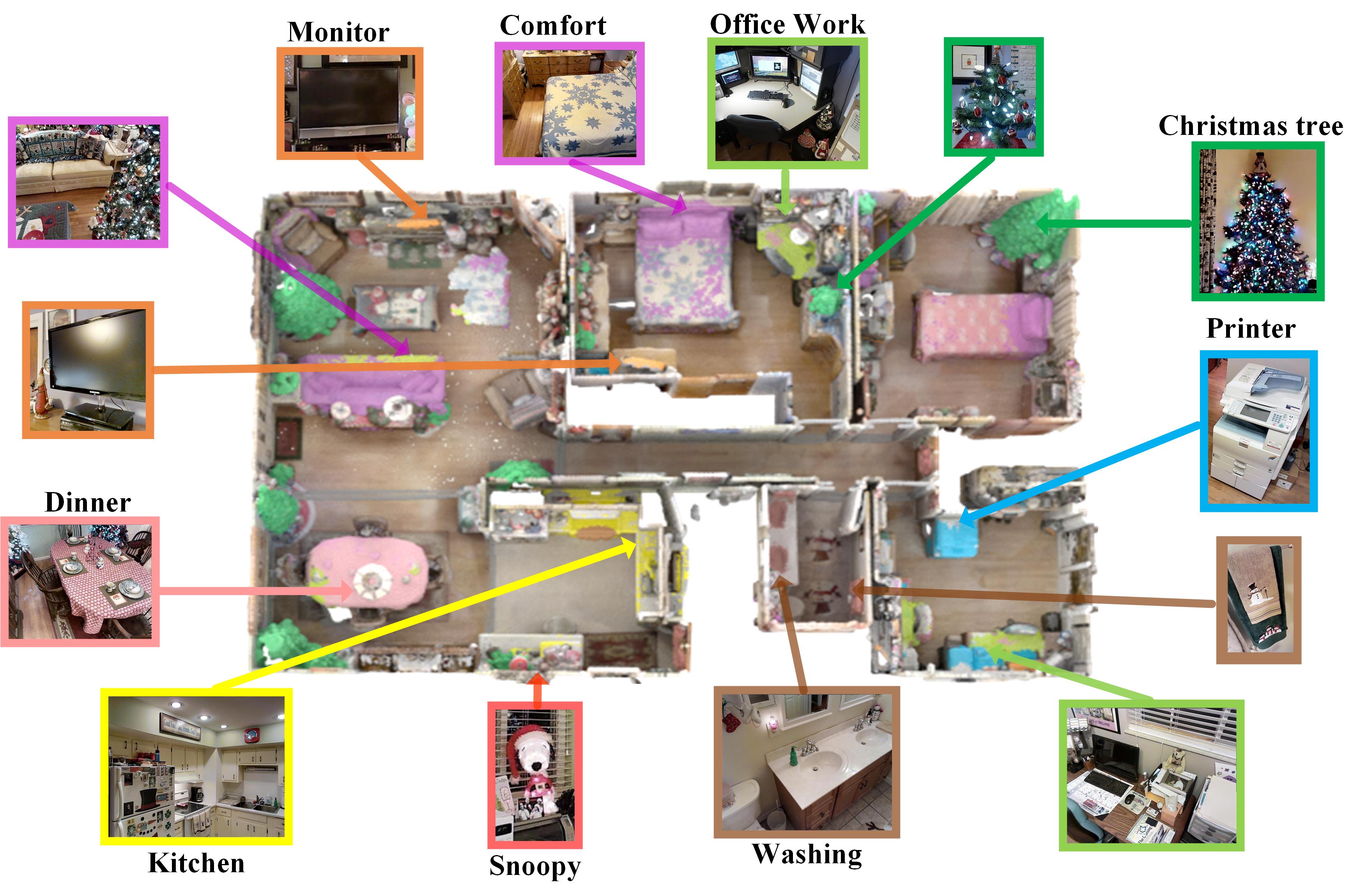}
	\vspace{-1.0em}
	\caption{Open-vocabulary segmentation results on rare categories and different forms of queries. The same color corresponds to the same query/category.}
	\vspace{-0.5cm}
	\label{DMA:rare_obj}
\end{figure}

%and the strong geometrical priors of 3D encoder can be incorporated into the 2D counterpart, maximizing the synergistic benefits of both modalities.  

%\begin{figure}[!t]
%	\centering 
%	\includegraphics[scale=0.3]{rare_obj_.jpg}\\ 
%	\vspace{-0.5em}
%	\caption{Open-vocabulary segmentation results on rare categories and different forms of queries. The same color corresponds to the same query/category.}
%	\label{DMA:rare_obj} 
%	\vspace{-0.5cm}
%\end{figure}

{\bf Open-Vocabulary Segmentation for Different Text Queries.} We finally investigate the ability of our method to segment rare categories. As shown in Fig.~\ref{DMA:rare_obj}, our method can accurately segment the corresponding regions for the given texts/queries in 3D scenes, even for unseen categories. For instance, our well-trained model can quickly locate the position of new categories such as ``Snoopy'', or functional areas such as ``kitchen'', \etc. 
On one hand, we align with 2D CLIP features that have been trained with a vast corpus of text. On the other hand, we construct a comprehensive and scalable textual modality by using VLMs, further enhancing the understanding ability.

%{\bf Tagging \vs Caption. } We compare the effects of image tagging and detailed scene descriptions in Tab.~\ref{tab:tagging}.  

%{\bf Qualitative results. }\\
%{\bf Non-mutual exclusive loss. }\\
%{\bf Distilling 3D priors to 2D. }\\

\section{Conclusion}
We presented a dense multimodal alignment (DMA) framework for open-vocabulary 3D scene understanding by establishing dense correspondences between 3D points, 2D images and 1D texts, and leveraging their synergistic benefits to learn robust and generalizable 3D representations. To build a scalable language modality, we utilized powerful vision-language models to extract comprehensive scene descriptions and category information. Furthermore, we preserved the open-vocabulary recognition ability of the image modality  by combining frozen CLIP features with trainable mask features. Extensive experiments demonstrate the promising performance of our method in open-vocabulary segmentation tasks across various indoor and outdoor scenarios.

\textbf{Limitations.} Our method relies on the quality of generated text descriptions and image features. In addition, collecting a larger 3D scene dataset is crucial for improving the generalization ability to unseen categories and variations.  

%In conclusion, our work introduces a dense multimodal alignment framework for open-vocabulary 3D scene understanding. We establish dense correspondences between 3D points, image pixels, and text labels, leveraging their synergistic benefits to learn robust and generalizable 3D representations. To generate a scalable language modality, we utilize powerful vision-language models to extract comprehensive scene descriptions and category information. Furthermore, we preserve the open-vocabulary recognition ability by combining frozen CLIP features with trainable mask features for the image modality. Extensive experiments demonstrate the promising performance of our method in open-vocabulary segmentation tasks across various indoor and outdoor scenarios. Our approach addresses the challenge of costly 3D data annotation and expands the application of 3D scene understanding methods to real-world settings with unlimited possible classes.

\bibliographystyle{splncs04}
\bibliography{main}

\clearpage 

In this supplemental file, we provide the following materials:
\begin{itemize}%[leftmargin=*]
	\item[$\bullet$] The instructions to reduce noisy tags via GPT and the corresponding visualizations of 3D label map by using denoised tags (referring to ``\textbf{Sec.~3.1}-Comprehensive Text Modality Generation-\textit{Reliable GPT-based Denoising}'' in the main paper); 
	\item[$\bullet$] Examples of scene-level captions extracted by MLLMs and the corresponding visualizations of 3D label map by using the captions as queries (referring to ``\textbf{Sec.~3.1}-Comprehensive Text Modality Generation-\textit{Scalable Scene Descriptions}'' in the main paper);
	\item[$\bullet$] The impacts of tagging models and MLLMs on different datasets (referring to ``\textbf{Sec.~3.1}-Comprehensive Text Modality Generation'' and ``\textbf{Sec.~4.3}-Ablation Study-\textit{tagging models \vs MLLMs}'' in the main paper);
	\item[$\bullet$] Effect of Mutually Inclusive Loss (\textbf{MIL}) (referring to ``\textbf{Sec.~3.4}-Dense Multimodal Alignment-\textit{Mutually Inclusive Loss}'' in the main paper);
	%	\item[$\bullet$] Implementation details (referring to ``\textbf{Sec~4}-Experiments'' in the main paper);
	\item[$\bullet$] More qualitative results on indoor and outdoor datasets (referring to ``\textbf{Sec.~4.2}-Comparison with State-of-The-Arts'' in the main paper).
\end{itemize}

\section{Reducing Noisy Tags}
\begin{figure*}[!h]
	\centering 
	\includegraphics[scale=0.5]{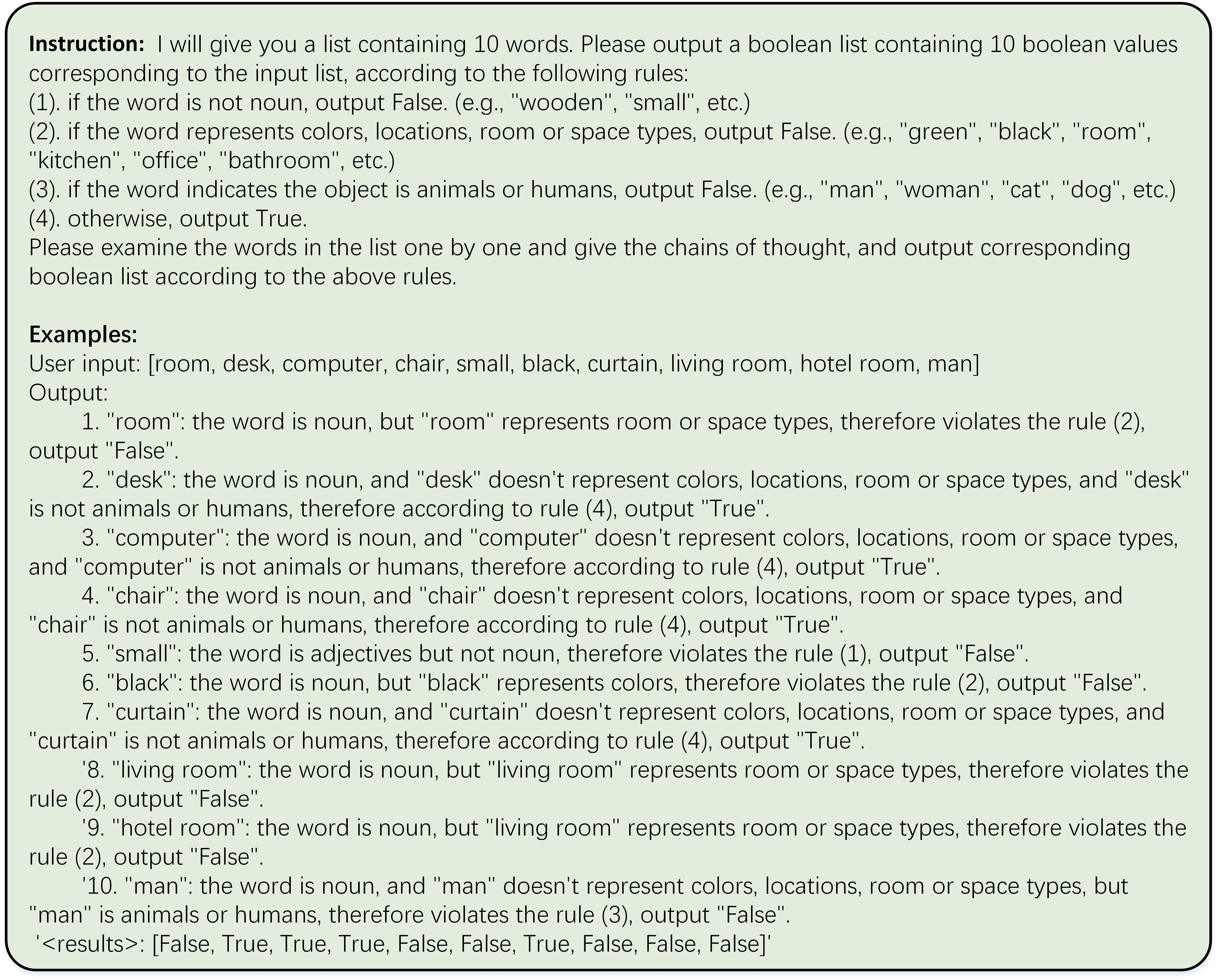}\\
	\caption{Instruction and examples to reduce noisy tags.}
	\label{DMA:chatbox}
\end{figure*}

\textbf{Instruction.} We use GPT to reduce noisy tags generated by RAM~\cite{zhang2023recognize}. The corresponding instructions and examples are given in Fig.~\ref{DMA:chatbox}. Given the input list of tags, we instruct GPT to evaluate each word individually to determine if it meets the given rules, and we ask it to output the chain of thought and the final boolean list.

\begin{figure*}
	\centering 
	\includegraphics[scale=0.4]{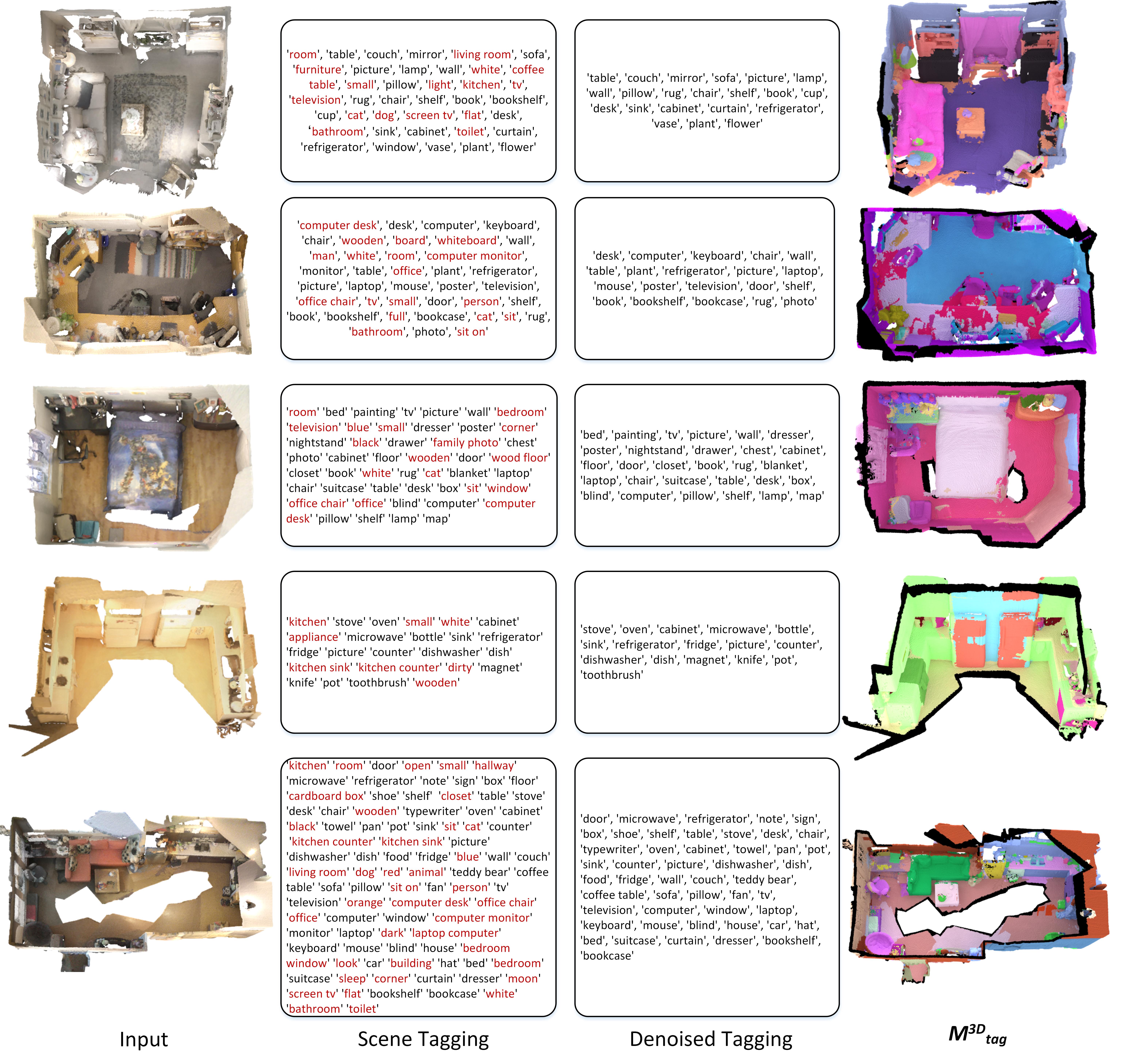}\\
	\caption{Examples of scene tagging results by RAM~\cite{zhang2023recognize} and visualizations of 3D label map $M^{3D}_{tag}$ by employing the tags as queries. The \textcolor{red}{red} words denote the noisy tags which are filtered out by GPT and multi-view voting.}
	\label{DMA:painted_scene}
\end{figure*}

\noindent \textbf{Denoised tagging results.} We give some examples of the denoised scene tags and the corresponding visualizations in Fig.~\ref{DMA:painted_scene}. We can observe that many noisy tags can be filtered out in terms of the rules given in Fig.~\ref{DMA:chatbox}. We also visualize the text-to-3D label map $M^{3D}_{tag}$ as defined in Eq.~2 of the main paper. Label errors in this 3D label map are consequently reduced based on the denoised tags.

\section{Scene Captions} 
\begin{figure*}
	\centering 
	\includegraphics[scale=0.5]{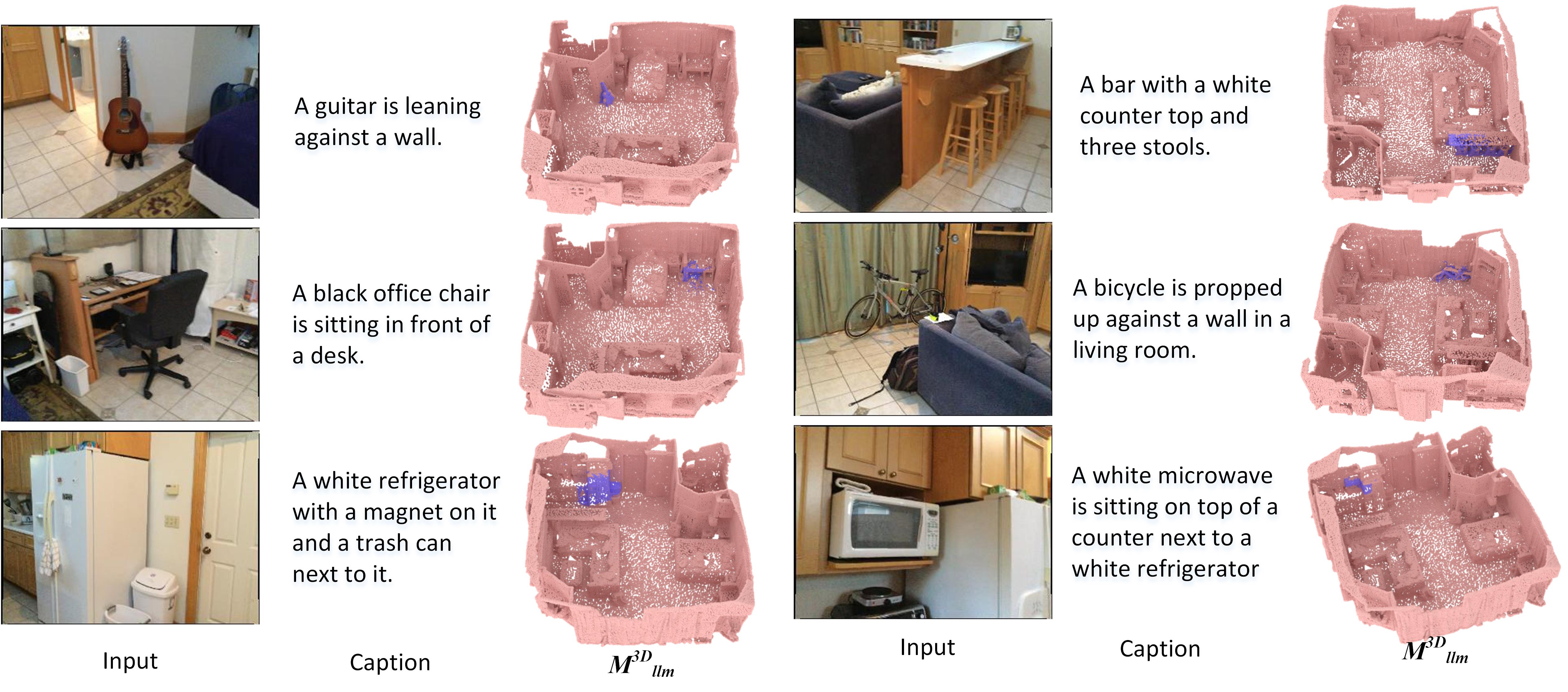}\\
	\caption{Examples of scene captions extracted by LLaVA~\cite{liu2023llava} and visualizations of 3D label map $M^{3D}_{llm}$ by employing the captions as queries. We can build dense text-to-3D correspondences based on the obtained captions.}
	\label{DMA:llm}
\end{figure*}

In addition to category names, we also leverage MLLMss to generate comprehensive descriptions for each view. Fig.~\ref{DMA:llm} shows some examples of the scene captions extracted by LLaVA~\cite{liu2023llava} and the visualizations of 3D label map $M^{3D}_{llm}$ generated by employing the captions as queries. We can build dense text(caption)-to-3D correspondences, thus achieving more accurate and fine-grained alignment.

\section{Tagging Models \vs MLLMs}
In Tab.~\ref{tab:tagging} we compare the results of aligning 3D points to image tags and captions on different datasets. The former involves complete category information, while the latter contains comprehensive scene descriptions and more contextual information. As can be seen, tagging model plays a crucial role in the overall performance, as it encompasses more extensive categories and semantics. MLLM further enhances the final performance by incorporating rich contextual semantics and diversified descriptions. It is worth noting that MLLM only contributes to a marginal performance improvement on the nuScenes~\cite{caesar2020nuscenes} dataset since this dataset contains a limited number of image views.  
\begin{table}[!h]
	\centering
	\scalebox{1.0}{
		\begin{tabular}{c|ccc}
			\toprule \rowcolor{cyan!15}
			& ScanNet & Matterport3D & nuScenes \\ \hline \hline 
			Tagging+GPT & 46.3    &   37.7           & 43.3     \\
			MLLM & 24.2    &   19.4           & 11.3     \\
			Both    & 50.5    &   39.8       & 43.8    \\ \bottomrule
	\end{tabular}}
	\caption{Performance comparison of tagging model and MLLM.}
	\label{tab:tagging}
\end{table}

\section{Mutually Inclusive Loss}
As discussed in Sec.~3.4 of the main paper, we use mutually inclusive loss to align 3D representations to text modality. In Tab.~\ref{tab:multual_exclusive}, we compare the performance by using mutually \textbf{inclusive} (BCE loss) and \textbf{exclusive} (CE loss) losses. Compared to binary cross-entropy loss, the cross-entropy loss leads to a performance degradation of 6.4\% mIoU because it causes conflicts between text labels with similar semantics, making it difficult for the model to converge and limiting the performance improvement.  
\begin{table}[!h]
	\centering
	\begin{tabular}{c|cc}
		\toprule \rowcolor{cyan!15}
		& Cross-entropy & Binary cross-entropy \\ \hline
		mIoU & 46.9          & 53.3                 \\ \bottomrule
	\end{tabular}
	\caption{Comparison of mutually inclusive and exclusive losses on ScanNet dataset.}
	\label{tab:multual_exclusive}
\end{table}

\section{Qualitative Results}
\begin{figure*}
	\centering 
	\includegraphics[scale=0.65]{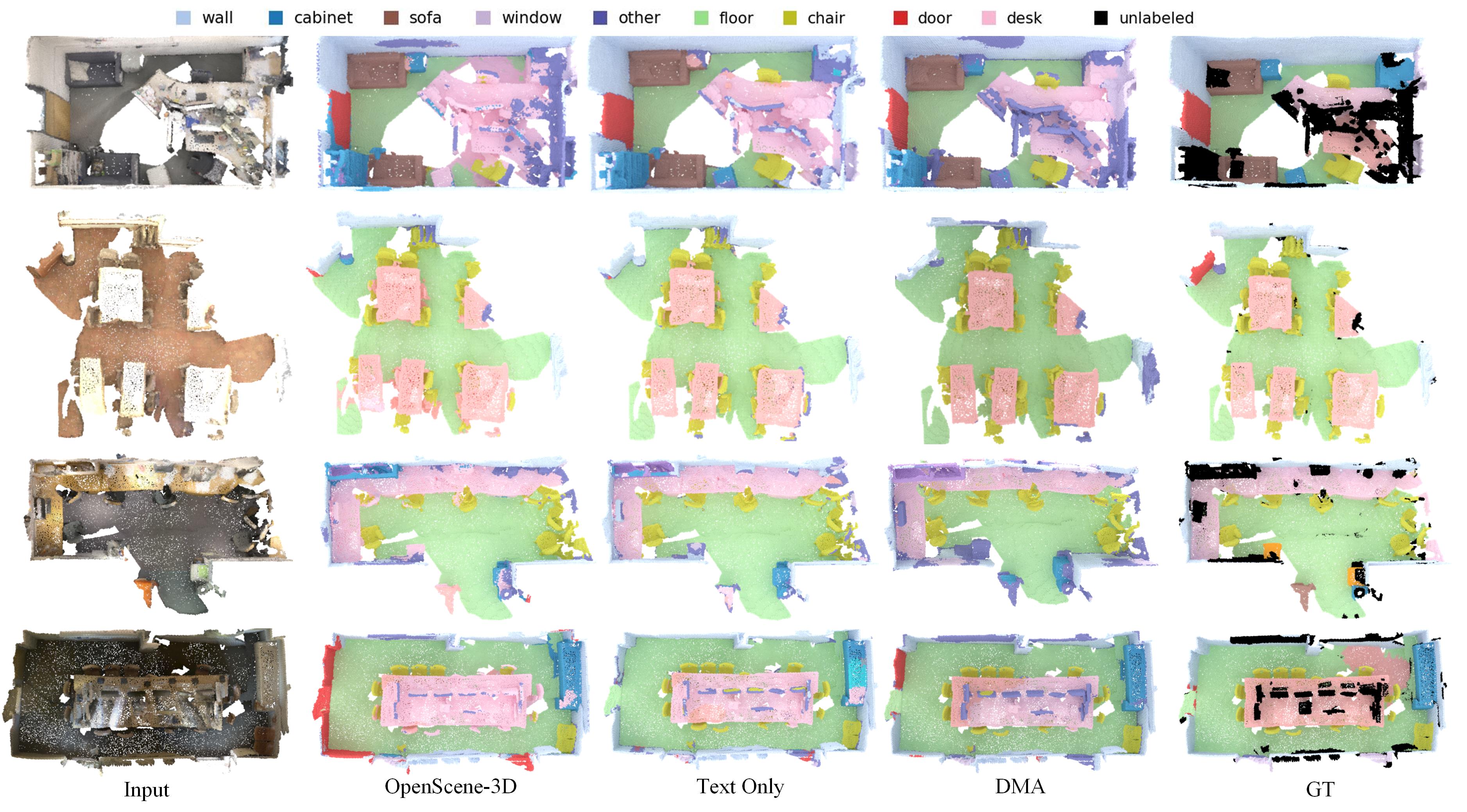}\\
	\caption{Qualitative results of different methods on ScanNet dataset.}
	\label{DMA:scannet}
\end{figure*}

\begin{figure*}
	\centering 
	\includegraphics[scale=1.0]{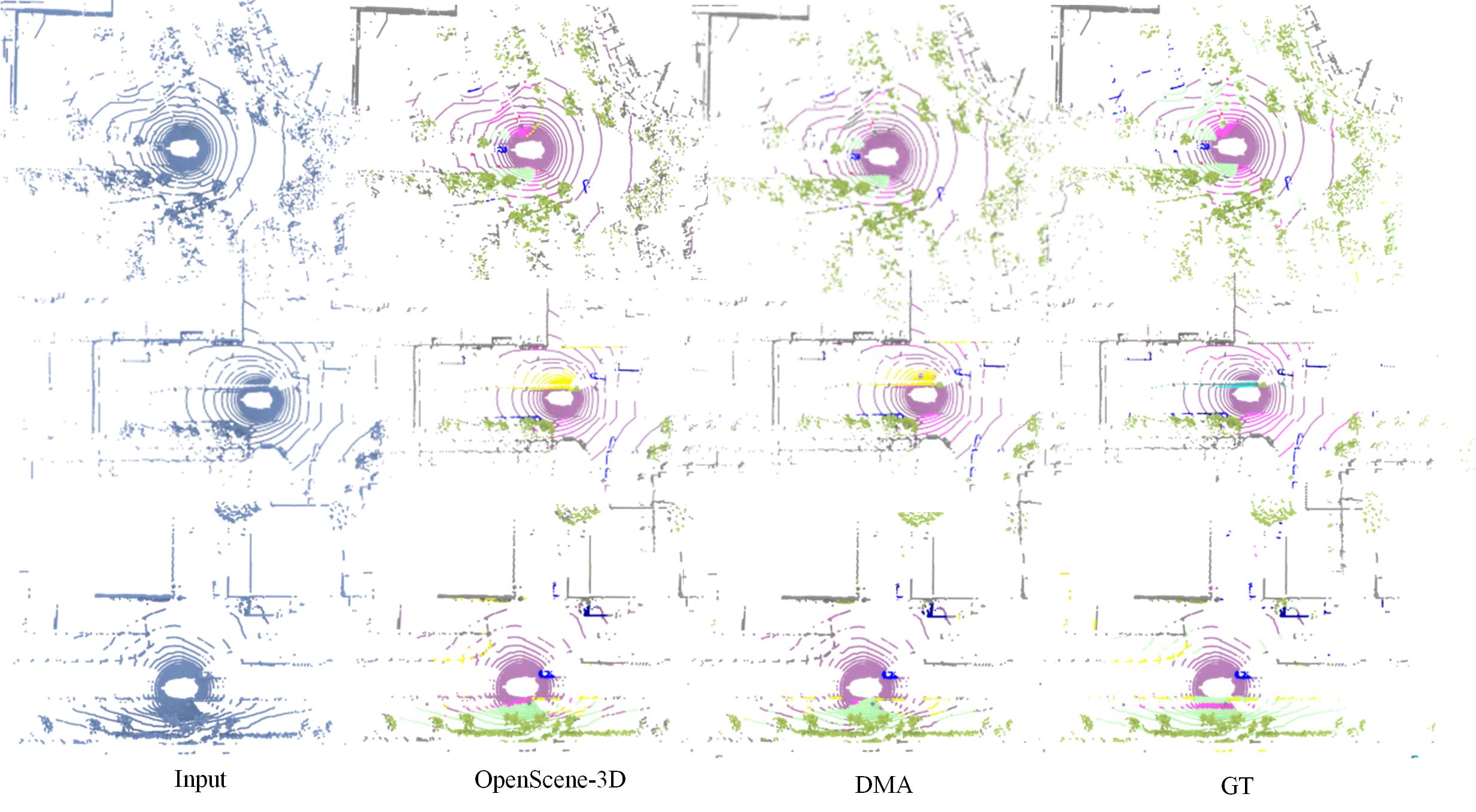}\\
	\caption{Qualitative results of different methods on nuScenes dataset.}
	\label{DMA:nuscenes}
\end{figure*}

We give more qualitative results on both indoor and outdoor datasets in Fig.~\ref{DMA:scannet} and Fig.~\ref{DMA:nuscenes}. Our method consistently produces better results on ScanNet~\cite{dai2017scannet} and nuScenes~\cite{caesar2020nuscenes} datasets, demonstrating the robustness and better generalization ability of the proposed method on different data distributions and open-set categories.

%\section{}
%Foreground categories: bed, chair, sofa, table, counter, desk, refrigerator, toilet, sink, bathtub, other furniture.
%
%\noindent Background categories: wall, floor, cabinet, door, window, bookshelf, picture, curtain, shower curtain.
%
%\noindent Base categories: car, drivable surface, other flat, sidewalk, terrain, man-made, vegetation.
%
%\noindent Long-tail categories: barrier, bicycle, bus, construction vehicle, motorcycle, pedestrian, traffic cone, trailer, truck.

\end{document}